\documentclass[journal]{IEEEtran}
\usepackage{cite}

\usepackage{multirow}
\usepackage{subfigure}
\usepackage{setspace}

\ifCLASSINFOpdf
  \usepackage[pdftex]{graphicx}
  \graphicspath{{../pdf/}{../jpeg/}}
  \DeclareGraphicsExtensions{.pdf,.jpeg,.png}
\else
\fi
\usepackage{amsmath}
\usepackage{amssymb}
\usepackage{algorithm}
\usepackage{algorithmic}

\usepackage{color}

\begin{document}

\title{Progressive Object Transfer Detection} 
%
%
%

\author{Hao~Chen$^{*}$,
        Yali~Wang$^{*}$,
        Guoyou~Wang, 
        Xiang~Bai, 
        and~Yu~Qiao$^{\dagger}$ 
\thanks{
This work is partially supported by 
National Key Research and Development Program of China (No. 2016YFC1400704),
National Natural Science Foundation of China (61876176,U1613211,U1713208), 
Shenzhen Basic Research Program (JCYJ20170818164704758), 
Joint Lab of CAS-HK.
}
\thanks{
$*$ H. Chen and Y. Wang are the \emph{equally-contributed first authors}.
}
\thanks{
$\dagger$ Y. Qiao is the \emph{corresponding author}.
}
\thanks{
H. Chen is with University of Maryland, College Park.
This work was done during his internship at Shenzhen Institutes of Advanced Technology, Chinese Academy of Sciences.
(Email: chenh@cs.umd.edu)
}
\thanks{
Y. Wang is with Shenzhen Key Lab. of C.V.P.R. and SIAT-Sensetime Joint Lab, Shenzhen Institutes of Advanced Technology, Chinese Academy of Sciences.
(Email: yl.wang@siat.ac.cn)
}
\thanks{
G. Wang is with Huazhong University of Science and Technology, China.
(Email: gywang@hust.edu.cn)
}
\thanks{
X. Bai is with Huazhong University of Science and Technology, China.
(Email: xbai@hust.edu.cn)
}
\thanks{
Y. Qiao is with Shenzhen Key Lab. of C.V.P.R. and SIAT-Sensetime Joint Lab, Shenzhen Institutes of Advanced Technology, Chinese Academy of Sciences.
(Email: yu.qiao@siat.ac.cn)
}
}

%
%

\markboth{IEEE TRANSACTIONS ON IMAGE PROCESSING}%
{Chen \MakeLowercase{\textit{et al.}}: IEEE TRANSACTIONS ON IMAGE PROCESSING}
%



\maketitle

\begin{abstract}
Recent development of object detection mainly depends on deep learning with large-scale benchmarks.
However,
collecting such fully-annotated data is often difficult or expensive for real-world applications,
which restricts the power of deep neural networks in practice.
Alternatively,
humans can detect new objects with little annotation burden,
since humans often use the prior knowledge to identify new objects with few elaborately-annotated examples,
and subsequently generalize this capacity by exploiting objects from wild images.
Inspired by this procedure of learning to detect,
we propose a novel Progressive Object Transfer Detection (POTD) framework.
Specifically,
we make three main contributions in this paper.
First,
POTD can leverage various object supervision of different domains effectively into a progressive detection procedure.
Via such human-like learning,
one can boost a target detection task with few annotations.
Second,
POTD consists of two delicate transfer stages,
i.e.,
Low-Shot Transfer Detection (LSTD),
and
Weakly-Supervised Transfer Detection (WSTD).
In LSTD,
we distill the implicit object knowledge of source detector to enhance target detector with few annotations.
It can effectively warm up WSTD later on.
In WSTD,
we design a recurrent object labelling mechanism for learning to annotate weakly-labeled images.
More importantly,
we exploit the reliable object supervision from LSTD,
which can further enhance the robustness of target detector in the WSTD stage.
Finally,
we perform extensive experiments on a number of challenging detection benchmarks with different settings.
The results demonstrate that,
our POTD outperforms the recent state-of-the-art approaches.
The codes and models are available at \emph{https://github.com/Cassie94/LSTD/tree/lstd}.

\end{abstract}



%
\IEEEpeerreviewmaketitle


\begin{figure*}[t]
\centering
\includegraphics[width=.95\textwidth]{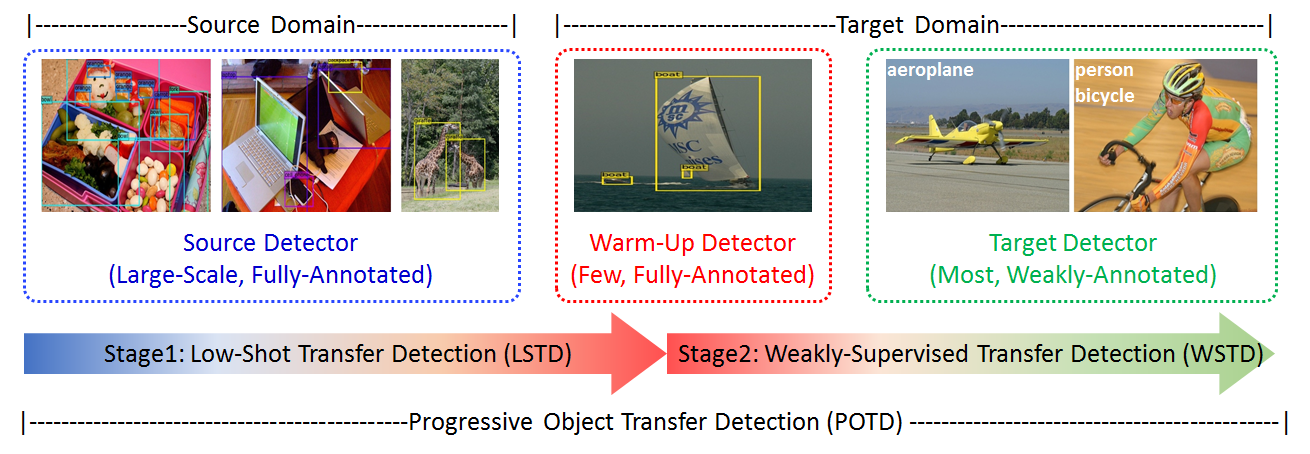}
\caption{
Progressive Object Transfer Detection (POTD).
It consists of two stages to imitate the learning procedure of humans.
\textbf{First},
we design a Low-Shot Transfer Detection (LSTD) stage to imitate the warm-up process,
i.e.,
humans use the prior object knowledge to check new objects with few elaborately-annotated images.
To achieve it,
we fine-tune source detector with few fully-annotated target images.
By distilling implicit source object knowledge,
we can effectively warm up the low-shot detection in the target domain.
\textbf{Second},
we design a Weakly-Supervised Transfer Detection (WSTD) stage to imitate the generalization process,
i.e.,
humans generalize the warm-up stage by exploiting objects from wild images without full annotations.
To achieve it,
we design an effective recurrent object labelling mechanism,
which can learn to detect with weakly-annotated images.
Furthermore,
we exploit the reliable object supervision from the LSTD stage,
in order to further generalize the detection performance in the target domain.
Via this human-like progressive learning,
POTD can boost detection with little annotation burden.
Better viewed in color.
}
\label{SSTDidea}
\end{figure*}

\section{Introduction}

Recent years have witnessed the breakthrough of object detection,
with the development of deep learning frameworks \cite{7410526, ren2015faster, Liu2016SSDSS, he2017mask}.
However,
the remarkable performance of these detectors heavily depends on the large-scale benchmarks with object annotations.
For a specific detection task in practice,
it is often labor-intensive to collect such fully-annotated data.
To alleviate this annotation burden,
a number of weakly/semi-supervised detectors have been proposed with weakly-annotated images (i.e., with only image labels) \cite{8100028, 7780751, 7780680, Lai2017SaliencyGE, 8099809, dong2017few}.
Even though such images can be obtained easily from the internet,
the performance of these weakly/semi-supervised detectors is far from being competitive to fully-supervised detectors.

Alternatively,
human can localize and recognize new objects successfully,
after checking few examples with the prior object knowledge.
Moreover,
this capacity can be generalized via exploiting objects from weakly-annotated images.
To mimic this learning process,
several approaches have been proposed recently,
by low-shot and/or transfer learning in a semi-supervised detection setting \cite{Hoffman2014LSDALS,7780602,chen2018lstd}.
However,
these detectors have difficulties in handling wild images with complex objects \cite{Hoffman2014LSDALS,7780602},
or learning to detect with weakly-annotated images \cite{chen2018lstd}.
For this reason,
\cite{Liang2015TowardsCB} introduces a baby learning framework,
based on prior knowledge modelling, exemplar learning, and learning with video contexts.
However,
it often requires a large amount of weakly-annotated videos (e.g., 20,000 per category),
and the iterative learning manner may reduce the discriminative power of deep neural networks.

To address these challenges,
we propose a novel Progressive Object Transfer Detection (POTD) framework in Fig. \ref{SSTDidea}.
Our key insight is that,
\textbf{
human-like learning can effectively integrate various object knowledge of different domains into a progressive detection process,
which can boost a target detection task with very few instance-level annotations.
}
More specifically,
this paper makes three main contributions.
\textbf{First},
POTD can efficiently mimic the learning manner of human,
by progressively transferring object knowledge from source to target, from large-scale to low-shot, from fully-annotated to weakly-annotated images.
With this multi-level learning process,
one can effectively promote the detection performance with little annotation burden.
\textbf{Second},
POTD consists of two novel detection stages,
i.e.,
Low-Shot Transfer Detection (LSTD)
and
Weakly-Supervised Transfer Detection (WSTD).
Each stage is elaborately designed with transfer insights.
In LSTD,
we distill implicit object knowledge in source to guide low-shot detection in target.
This can effectively warm up WSTD to handle wild images later on.
In WSTD,
we design a recurrent object labelling mechanism for learning to detect with weakly-labeled images.
Furthermore,
we exploit reliable object knowledge from LSTD,
which can enhance the robustness of target detector for weakly-supervised detection.
\textbf{Finally},
we conduct extensive experiments on a number of challenging data sets,
where
our POTD outperforms the recent state-of-the-art approaches.

It is noted that,
Low-Shot Transfer Detection (LSTD) has been published in \textbf{AAAI 2018} \cite{chen2018lstd}.
We significantly extend it in the following ways.
From the aspect of model designs,
we extend LSTD to be a Progressive Object Transfer Detection (POTD) procedure.
Specifically,
we use LSTD as warm-up,
and design a new Weakly-Supervised Transfer Detection (WSTD) stage to handle weakly-annotated images in the target domain.
Consequently,
our POTD can further boost the detection performance of LSTD with little annotation burden.
From the aspect of experiments,
we deeply investigate the proposed POTD,
and show the effectiveness of both LSTD and WSTD on object detection.

\section{Related Work}

\textbf{Weakly-Supervised Object Detection}.
Since only image labels are available in the weakly-supervised object detection,
most approaches are naturally based on the Multiple Instance Learning (MIL) framework \cite{Dietterich1997SolvingTM},
where each image is regarded as a bag of object instances.
Positive categories are assumed to contain at least one object instance in the image,
while negative categories contain no corresponding instances at all.
However,
MIL alternates between estimating the object representation and selecting positive object instances,
which often converges to an unsatisfied local optima.
To alleviate this problem,
a number of efforts have been made by
good initializations \cite{7420739, 7780751},
suitable optimization strategies\cite{7298711,8099940},
and so on.
Recently,
deep learning has been used for weakly-supervised detection \cite{7780680,8100028,Lai2017SaliencyGE,8099809}.
A fundamental framework refers to weakly supervised deep detection network \cite{7780680},
which performs localization and classification within a two-stream architecture.
To further improve performance,
several extensions have been introduced by
context design \cite{Kantorov2016ContextLocNetCD},
saliency guidance \cite{Lai2017SaliencyGE},
online classifier refinement \cite{8099809},
etc.
However,
the unsupervised selective search \cite{Uijlings2013SelectiveSF} or edgeBoxes\cite{Zitnick2014EdgeBL} may limit the efficiency of these deep detection networks.
More importantly,
the performance of these detectors is considerably lower than that of fully-supervised detectors,
due to the lack of elaborate object annotations.

\textbf{Semi-Supervised Object Detection}.
Most semi-supervised detectors assume that several categories are fully-annotated \cite{Hoffman2014LSDALS, 7780602}.
Via transferring the detection knowledge of these categories into the weakly-annotated ones,
these detectors can improve the final performance on the whole set.
However,
\cite{Hoffman2014LSDALS, 7780602} attempt to adapt an image classifier to an object classifier,
which is often effective on single-instance images.
It may limit the detection capacity for wild images.
In addition,
these detectors often require moderate amount of object annotations,
which can be still difficult or expensive to obtain.

\textbf{Low-Shot Transfer Object Detection}.
Alternatively,
\cite{dong2017few} proposes a more reasonable data assumption to alleviate labelling burden,
i.e.,
quite a few images are fully-annotated for each object category,
and all other images are weakly-annotated with only image labels.
But its performance is limited,
without guidance of prior object knowledge.
For this reason,
several low-shot and/or transfer detection approaches \cite{Liang2015TowardsCB, chen2018lstd} have been proposed recently,
in order to mimic the learning procedure of human.
By taking advantage of large-scale benchmarks in source,
these detectors can work with few object annotations in target.
However,
\cite{chen2018lstd} ignores the weakly-annotated target images,
which may restrict the generalization capacity of target detector.
\cite{Liang2015TowardsCB} attempts to borrow numerous weakly-annotated videos,
but the iterative learning scheme lacks the efficiency of end-to-end learning.
Different from these approaches,
our POTD is a progressive learning procedure,
which is built upon the fully end-to-end training framework with very few object annotations.

\begin{figure*}[t]
\centering
\includegraphics[width=.95\textwidth]{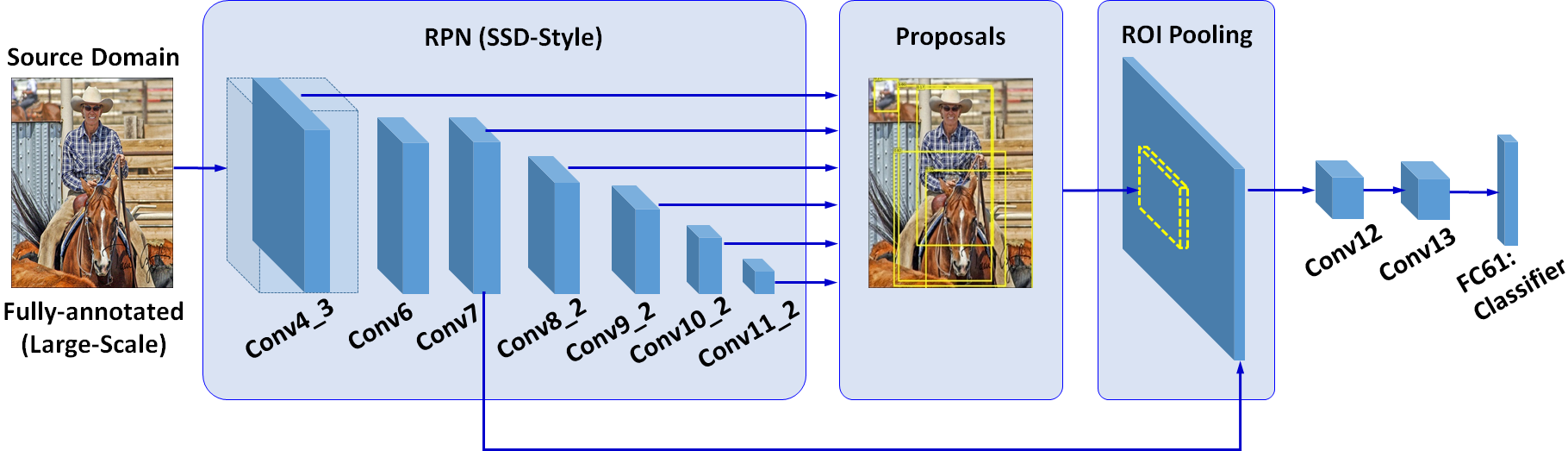}
\caption{Basic Deep Architecture.
It is a Faster-RCNN-style detection framework but with SSD-style region proposal network (RPN).
By leveraging the core designs in both detectors,
our framework can alleviate the transfer difficulties from large-scale source domain to low-shot target domain.
More details can be found in Section \ref{Basic Deep Architecture of LSTD}.
Better viewed in color.}
\label{SourceDetector}
\end{figure*}

\section{Overall Learning Procedure of Progressive Object Transfer Detection (POTD)}

\textbf{Problem Definition.}
To effectively address a target detection task with little annotation burden,
we define our detection problem according to the following settings in practice.
\textbf{First},
we can access a well-trained detector in the source domain.
\textbf{Second},
we partially annotate the training set in the target domain,
e.g.,
quite a few images (like one-shot per category) are fully-annotated with object bounding boxes,
and all others are weakly-annotated with only image labels.
\textbf{Finally},
we consider one of the most challenging transfer cases,
where the object categories in source and target are non-overlapped.

\textbf{Overall Learning Procedure.}
In this work,
we design a novel Progressive Object Transfer Detection (POTD) framework,
for learning to detect like humans.
The whole procedure of POTD is shown in Fig. \ref{SSTDidea},
which consists of two elegant transfer stages,
i.e.,
Low-Shot Transfer Detection (LSTD),
and
Weakly-Supervised Transfer Detection (WSTD).
In \textbf{LSTD},
we adapt the source-domain detector to be a warm-up detector in the target domain.
This stage can stabilize WSTD afterwards,
by distilling rich object knowledge in the source domain.
In \textbf{WSTD},
we adapt the warm-up detector to be a target detector,
which can learn to detect with weakly-annotated images in a fully end-to-end learning manner.
Via this human-like learning procedure,
our POTD can effectively address the target detection task with little annotation burden.

\textbf{Notations.}
To be concise,
we list all the relevant notations in the procedure of POTD,
w.r.t.,
two transfer detection stages including LSTD and WSTD in Table \ref{listnote}.
In the following,
we describe these two stages of POTD in detail.

\begin{table}[t]
\centering
\caption{Notations of Progressive Object Transfer Detection (POTD).}
\begin{spacing}{1.3}
\resizebox{0.45\textwidth}{!}{
\begin{tabular}{ll}
\hline
\textbf{General} & \textbf{Description} \\
\hline
$C_{s}$ & number of object classes in the source domain   \\
$C_{t}$ &  number of object classes in the target domain   \\
$K$ &  number of object proposals \\
\hline
\textbf{LSTD} & \textbf{Description }   \\
\hline
$\mathcal{L}_{lstd}$ &  total training loss                 \\
$\mathcal{L}_{lstd}^{main}$ & main loss (i.e., standard detection loss)       \\
$\lambda_{lstd}^{main}$ & coefficient of main loss \\
$\mathcal{L}_{lstd}^{bd}$   &   loss of background depression (BD)\\
$\lambda_{lstd}^{bd}$ &  coefficient of BD loss               \\
$\mathbf{F}^{bd}$    & feature regions that refer to image background\\
$\mathcal{L}_{lstd}^{sdk}$   &   loss of source detection knowledge (SDK)\\
$\lambda_{lstd}^{sdk}$ &  coefficient of SDK loss               \\
$\mathbf{Q}^{sdk}$    & SDK in source detector\\
$\mathbf{S}^{sdk}$    & SDK in warm-up detector  \\
\hline
\textbf{WSTD} & \textbf{Description}  \\
\hline
$\mathcal{L}_{wstd}$ & total training loss             \\
$\mathcal{L}_{wstd}^{sdk}$ & SDK loss    \\
$\lambda_{wstd}^{sdk}$    &  coefficient of SDK loss  \\
$\mathbf{P}^{sdk}$ & SDK in target detector \\
$\mathcal{L}_{wstd}^{rol}$ & loss of recurrent object labelling (ROL) \\
$\lambda_{wstd}^{rol}$ & coefficient of ROL loss \\
$\mathcal{L}^{i}_{rol}$  & ROL loss for Classifier $i$ \\
$\mathbf{P}^{img}$ & Prediction score vector of one image  \\
$\mathbf{Y}^{img}$ & Ground truth image label\\
$\mathbf{P}^{i}$ & Prediction score matrix of proposals for Classifier $i$\\
$\mathbf{S}^{i}$  & Pseudo label matrix of proposals for Classifier $i$\\
\hline
\end{tabular}
}
\end{spacing}
\label{listnote}
\end{table}

%

\section{Low-Shot Transfer Detection (LSTD)}

Since detection with weak annotations often gets stuck in local optima,
we first design a warm-up stage,
i.e.,
fine-tuning source-domain detector with fully-annotated images in the target domain.
However,
such annotated images are quite few (e.g., one-shot per category) for a target detection task,
in order to reduce the annotation burden.
This fact can significantly increase the fine-tuning difficulties.
For this reason,
we propose a novel low-shot transfer detection framework to alleviate overfitting.

\begin{figure*}[t]
\centering
\includegraphics[width=0.9\textwidth]{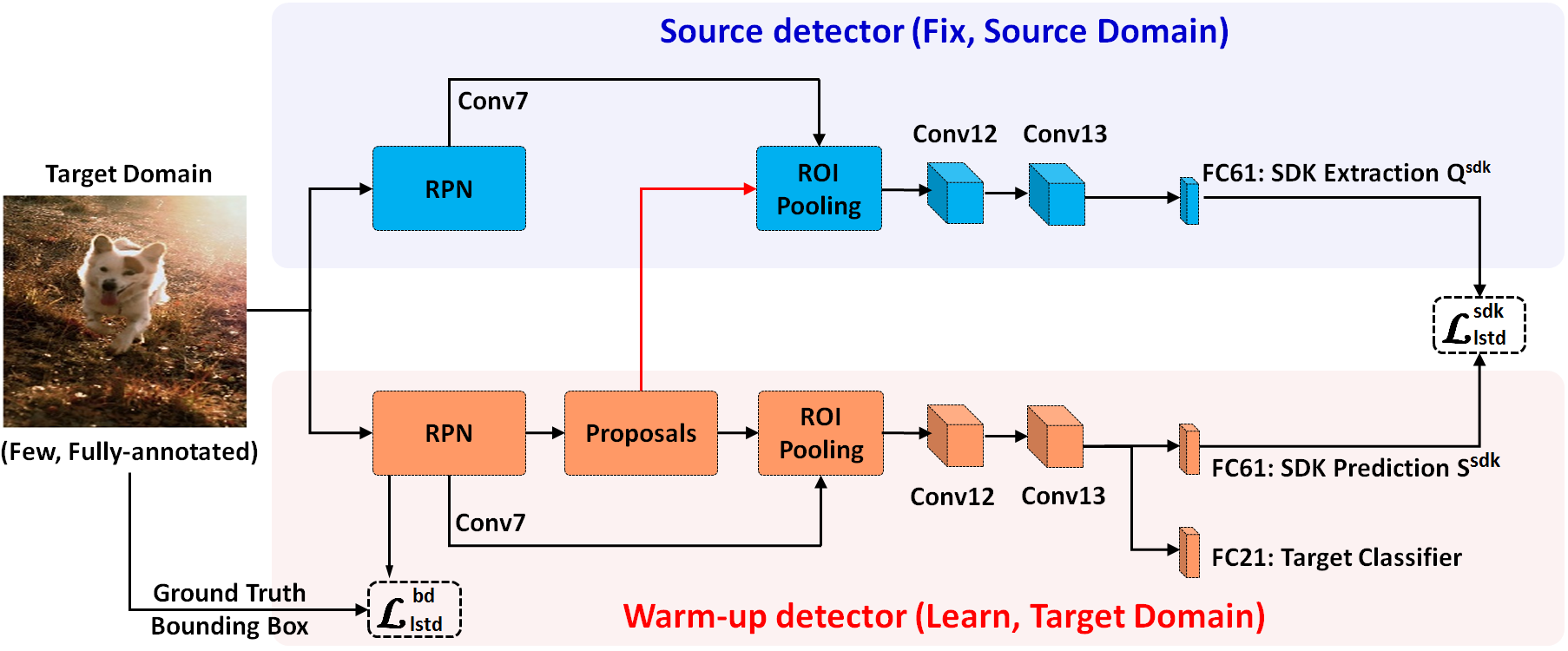}
\caption{Regularized Transfer Learning for LSTD.
First,
we pretrain the basic architecture (Fig. \ref{SourceDetector}),
by using the large-scale detection benchmark in the source domain.
The resulting source detector contains rich object knowledge for feature generalization.
Second,
we use source detector to initialize a warm-up detector,
and fine-tune it with the given fully-annotated training images in the target domain.
Due to low-shot scenario,
we propose two novel regularization terms during fine-tuning,
i.e.,
$\mathcal{L}_{lstd}^{bd}$ for background depression (BD),
and
$\mathcal{L}_{lstd}^{sdk}$ for source detection knowledge (SDK).
More details can be found in Section \ref{LSTD with Source Detection Knowledge (SDK)}.
Better viewed in color.}
\label{warm-up}
\end{figure*}

\subsection{Deep Detection Architecture for LSTD}
\label{Basic Deep Architecture of LSTD}
We first design a flexible deep learning architecture,
which can alleviate transfer difficulties from large-scale source domain to low-shot target domain.
As shown in Fig. \ref{SourceDetector},
it is a Faster-RCNN-style detection framework but with SSD-style region proposal network (RPN).

\textbf{Why to Choose Faster-RCNN-Style Framework.}
This is mainly credited to one key design in Faster RCNN,
i.e.,
region proposal network (RPN).
More specifically,
the object classifier in RPN is used to identify whether the proposal is object or not.
Hence,
it can learn the common traits of different objects,
e.g.,
clear edge,
uniform texture,
and so on.
By pretraining RPN with large-scale detection benchmark in the source domain,
we can obtain the high-quality proposals for low-shot detection in the target domain.
On the contrary,
the object classification in SSD is one-stage,
i.e.,
it has to deal with thousands of randomly-initialized proposals directly.
This fact can deteriorate the target detection task,
especially with only few annotated images.

\textbf{Why to Design SSD-Style RPN.}
In the standard RPN \cite{Renpami2016},
the bounding box regressor is designed separately for each categories.
It means that,
the parameters of regressor have to be re-initialized for each new domain,
since object categories in different domains are often non-overlapped in practice.
This often increases the overfitting risk during fine-tuning,
when the target domain only contains few annotated images.
Alternatively,
we adapt RPN into a SSD style.
Since the regressor in SSD is shared among all object categories,
the pretrained parameters of SSD-style RPN can be reused as initialization in the low-shot target domain.
This avoids the random re-initialization in the standard RPN,
and thus reduces the fine-tuning burdens in the low-shot domain.
In addition,
we directly use SSD-style RPN for object localization,
without regression refining after the ROI pooling in the standard Faster RCNN.
The main reason is that,
the multiple-convolutional-feature design of SSD is sufficient to localize objects with various sizes.

\textbf{Discussion.}
Our deep architecture aims at reducing transfer learning difficulties for low-shot detection in the target domain.
To achieve it,
we flexibly leverage the core designs of both Faster RCNN and SSD.
Additionally,
this detector performs bounding box regression and object classification on two relatively separate places,
which can further decompose the learning difficulties in low-shot detection.

\subsection{Regularized Transfer Learning for LSTD}
\label{LSTD with Source Detection Knowledge (SDK)}

After designing a flexible deep architecture,
we introduce an end-to-end regularized transfer learning framework for LSTD in Fig. \ref{warm-up}.
Specifically,
this transfer procedure involves two detectors.
\textbf{Source Detector}.
We first use the large-scale detection benchmark in the source domain to train the basic architecture in Fig. \ref{SourceDetector}.
As a result,
we obtain a source detector which contains rich source-domain object knowledge for generalization.
\textbf{Warm-Up Detector}.
After pretraining,
we use source detector as initialization,
and perform fine-tuning with a few fully-annotated training images in the target domain.
The resulting detector is a target-domain detector.
Since it can stabilize weakly-supervised transfer detection later on,
we call it as the warm-up detector.
We next describe how to perform fine-tuning to obtain the warm-up detector via LSTD.

\textbf{Total Loss for LSTD}.
Due to the low-shot property,
direct fine-tuning often traps into the overfitting risk.
To alleviate this challenge,
we propose two novel regularization terms,
i.e.,
background depression (BD) and source-detection knowledge (SDK).
Specifically,
the total loss of fine-tuning can be written as
\begin{equation}
\mathcal{L}_{lstd}=\lambda_{lstd}^{main}\mathcal{L}_{lstd}^{main}+\lambda_{lstd}^{bd}\mathcal{L}_{lstd}^{bd}+\lambda_{lstd}^{sdk}\mathcal{L}_{lstd}^{sdk},
\label{eq:totalloss}
\end{equation}
where
$\mathcal{L}_{lstd}^{main}$ refers to the standard loss summation of bounding box regression and object classification in the warm-up detector,
$\mathcal{L}_{lstd}^{bd}$ and $\mathcal{L}_{lstd}^{sdk}$ refer to BD and SDK regularization.
$\lambda_{lstd}^{main}$, $\lambda_{lstd}^{bd}$ and $\lambda_{lstd}^{sdk}$ are respectively the coefficients of the main loss, BD and SDK regularization.

\begin{figure}[t]
\centering
\includegraphics[width= 0.44\textwidth]{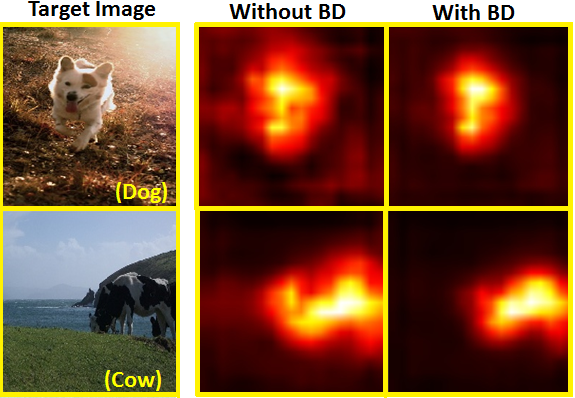}
\caption{BD regularization.
The feature map is obtained by averaging the convolutional feature cube (conv5$_{-}$3) over feature channels.
BD can successfully alleviate background disturbances on the feature map,
and thus allow warm-up detector to focus on target objects.}
\label{BDRegularization}
\end{figure}

\textbf{Background Depression (BD) Regularization}.
In the low-shot scenario,
the complex background may disturb the localization performance.
For this reason,
we propose a novel background-depression (BD) regularization,
by using the ground-truth bounding boxes of fully-annotated training images in the target domain.
First,
we feed a target image into warm-up detector,
and generate the convolutional feature cube from a middle-level convolutional layer.
Second,
we mask this convolutional cube with the ground-truth bounding boxes of all the objects in the image.
Consequently,
we can identify the feature regions that are corresponding to image background,
namely $\mathbf{F}^{bd}$.
To depress the background disturbances,
we use L2 regularization to penalize the activation of $\mathbf{F}^{bd}$,
\begin{equation}
\mathcal{L}_{lstd}^{bd}=\|\mathbf{F}^{bd}\|_{2}.
\label{eq:BDregularization}
\end{equation}
With this $\mathcal{L}_{lstd}^{bd}$,
warm-up detector can suppress background regions while pay more attention to target objects,
which is especially important for training with a few images.
As shown in Fig. \ref{BDRegularization},
our BD regularization can successfully reduce the background disturbances.

\begin{figure}[t]
\centering
\includegraphics[width= 0.48\textwidth]{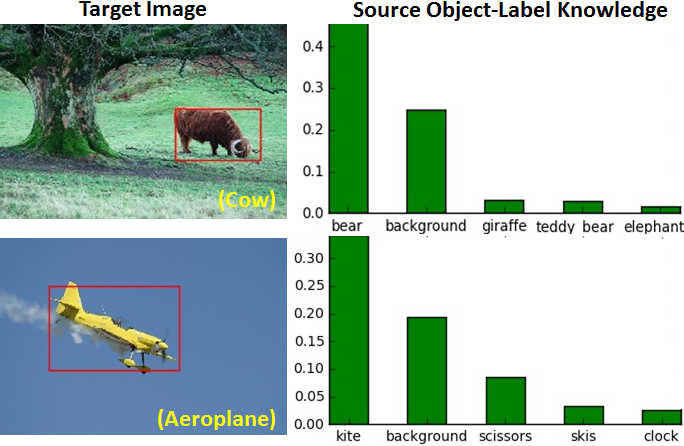}
\caption{SDK Visualization.
For a target-object proposal (red box: the proposal with the highest score),
we plot the top 5 probability scores of source objects.
One can see that,
the target object (\textit{Cow} or \textit{Aeroplane}) is relevant to (\textit{Bear} or \textit{Kite}) in the source domain,
due to the (color or shape) similarity.}
\label{TKRegularization}
\end{figure}

\textbf{Source Detection Knowledge (SDK) Regularization}.
Our key insight is that,
rich object knowledge in the large-scale source domain provides extra information about target domains.
As shown in Fig. \ref{TKRegularization},
\textit{Cow} or \textit{Aeroplane} may have a high response on \textit{Bear} or \textit{Kite},
due to the color or shape similarity.
Apparently,
this knowledge is important for fine-tuning warm-up detector,
especially with few fully-annotated images in the target domain.
Hence,
we propose to distill it for each object proposal in the target domain.

\textbf{(1) Extracting SDK from Source Detector}.
First,
we feed a target image into warm-up detector,
which can produce $K$ object proposals of this image.
Then,
we put these target proposals into the ROI pooling layer of source detector,
which can generate a SDK matrix $\mathbf{Q}^{sdk}\in\mathbb{R}^{(C_{s}+1)\times K}$ from the final FC layer.
Each column of $\mathbf{Q}^{sdk}$ is the probability vector of a target proposal,
w.r.t.,
$C_{s}+1$ categories (objects + background) in the source domain.

\textbf{(2) Leveraging SDK into Warm-Up Detector}.
To incorporate SDK into fine-tuning,
we add a SDK prediction branch at the end of warm-up detector.
This branch can produce a prediction matrix $\mathbf{S}^{sdk}\in\mathbb{R}^{(C_{s}+1)\times K}$ for $K$ target proposals,
where
each column is the prediction vector of a target proposal,
w.r.t.,
$C_{s}+1$ categories (objects + background) in the source domain.
Consequently,
we apply cross entropy between $\mathbf{Q}^{sdk}$ and $\mathbf{S}^{sdk}$ as a regularization,
\begin{equation}
\mathcal{L}_{lstd}^{sdk}=\sum\nolimits_{k=1}^{K}\sum\nolimits_{c=1}^{C_{s}+1}\mathbf{Q}^{sdk}(c,k)\log{\mathbf{S}^{sdk}(c,k)}.
\label{warmloss}
\end{equation}
In this case,
SDK can be effectively integrated into fine-tuning,
which generalizes low-shot detection in the target domain.


\textbf{(3) Discussion}.
LSTD is a learning step,
instead of a detector.
In this step,
we transfer source detector (large-scale, source domain) into warm-up detector (low-shot, target domain).
Specifically,
we train warm-up detector with the regularization of source detector,
i.e.,
we use source detector to extract SDK,
and leverage it as extra supervision for training warm-up detector.
As a result,
LSTD can generalize warm-up detector with low-shot training images in the target domain.
Moreover,
it is worth mentioning that,
our SDK transfer is different from knowledge distillation \cite{hinton2015distilling}.
First,
knowledge distillation is originally designed for model compression,
while our SDK is proposed for transfer learning.
Second,
SDK is performed on each proposal for object detection.
Hence,
it is not the standard way of knowledge transfer,
which often works on the whole image for object classification.

\begin{figure*}[t]
\centering
\includegraphics[width= 0.95\textwidth]{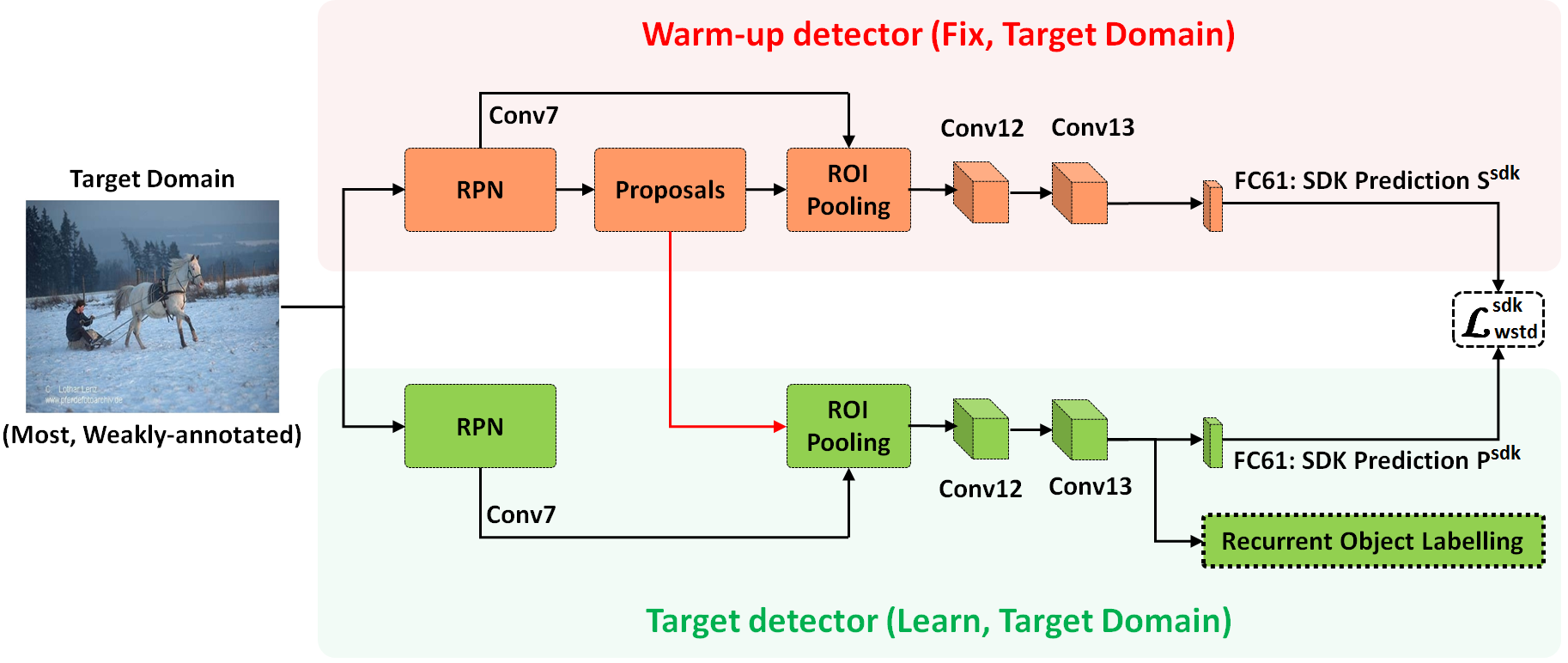}
\caption{Weakly Supervised Transfer Detection (WSTD).
On one hand,
we transfer object supervision of warm-up detector by $\mathcal{L}_{wstd}^{sdk}$,
which can enhance the generalization capacity of target detector.
On the other hand,
we design a novel recurrent object labelling (ROL) mechanism,
which can exploit confident proposals for learning to detect with weakly-annotated images in the target domain.
More details can be found in Section \ref{Weakly-Supervised Transfer Detection (WSTD)}.
Better viewed in color.}
\label{wstdshow}
\end{figure*}

\begin{figure*}[t]
\centering
\includegraphics[width= 0.98\textwidth]{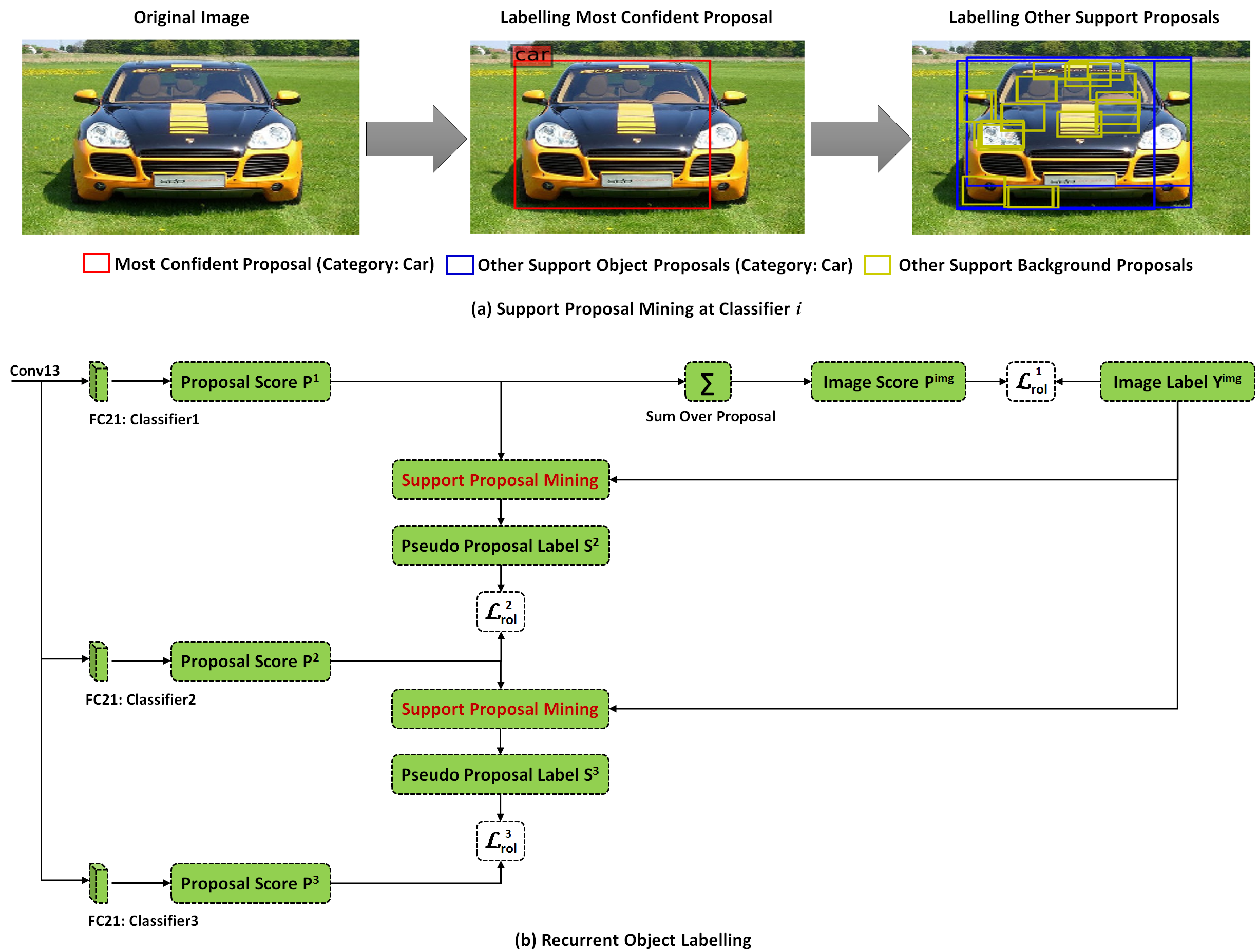}
\caption{Recurrent Object Labelling (ROL).
After generating proposals of a weakly-annotated image,
we feed them into ROI pooling of target detector in Fig. \ref{wstdshow}.
For each proposal,
conv13 in the target detector can generate a convolutional feature cube.
We subsequently feed these cubes into ROL.
\textbf{Classifier $i = 1$: Image-Level Supervision}.
We first pass all the convolutional cubes through Classifier $i=1$,
and obtain a prediction score matrix $\mathbf{P}^{1}$.
Then,
we sum $\mathbf{P}^{1}$ over proposals to generate an image score vector $\mathbf{P}^{img}$.
Finally,
we train Classifier $i = 1$ via Eq. (\ref{eq:LS1}),
i.e.,
multi-label loss between $\mathbf{P}^{img}$ and $\mathbf{Y}^{img}$ (i.e., ground truth image label).
\textbf{Classifier $i>1$: Support Proposal Mining}.
Since the training images are annotated with only image-level labels,
we propose to exploit object-level annotations as supervision.
Note that,
not all the proposals are discriminative.
Hence,
we design a support proposal mining mechanism in Subplot (a).
It first finds the most confident object proposal,
and then exploits other object and background proposals via IoU overlap.
Subsequently,
we use the generated pseudo label matrix $\mathbf{S}^{i}$ as supervision,
and train Classifier $i$ by cross entropy between $\mathbf{S}^{i}$ and $\mathbf{P}^{i}$.
To sum up,
ROL allows target detector to exploit the confident instance-level supervision in an online and recurrent manner.
This can further boost the detection performance with weakly-annotated images.
More details can be found in Section \ref{Recurrent Object Labelling (ROL)}.
Better viewed in color.}
\label{rolvisualize}
\end{figure*}

\section{Weakly-Supervised Transfer Detection (WSTD)}
\label{Weakly-Supervised Transfer Detection (WSTD)}

Via LSTD,
we transfer source detector into warm-up detector in the target domain.
Next,
we apply weakly-annotated images to further boost target detection task.
To achieve it,
we design another critical stage of POTD,
i.e.,
Weakly-Supervised Transfer Detection (WSTD),
which can effectively handle weakly-annotated images in a fully end-to-end transfer framework (Fig. \ref{wstdshow}).
Specifically,
this transfer procedure involves two detectors.
\textbf{Warm-Up Detector}.
LSTD produces a warm-up detector,
which is transferred from source detector, and trained on fully-annotated target images.
Hence,
this detector leverages the object knowledge from both source and target domains.
\textbf{Target Detector}.
We use warm-up detector as initialization,
and perform fine-tuning with weakly-annotated images in the target domain.
The resulting detector is a target-domain detector.
Since it is used to produce the final detection result in the target domain,
we call it as the target detector.
We next describe how to perform fine-tuning to obtain target detector via WSTD.

\textbf{Total Loss for WSTD}.
Specifically,
we introduce the total loss of WSTD with two distinct parts,
\begin{equation}
\mathcal{L}_{wstd}=\lambda_{wstd}^{sdk}\mathcal{L}_{wstd}^{sdk}+\lambda_{wstd}^{rol}\mathcal{L}_{wstd}^{rol}.
\label{Eq:TKfinetune}
\end{equation}
On one hand,
we transfer object supervision of warm-up detector via $\mathcal{L}_{wstd}^{sdk}$,
which can further regularize target detector to enhance the generalization capacity.
On the other hand,
we design a novel recurrent object labelling (ROL) mechanism via $\mathcal{L}_{wstd}^{rol}$,
which can exploit confident proposals for learning to detect with weakly-annotated images in the target domain.

\subsection{Object Supervision From Warm-Up Detector}

As mentioned before,
weakly-annotated images lack object-level supervision,
which makes the learning procedure trap into an unsatisfactory solution.
Alternatively,
warm-up detector can inherit rich object knowledge of source detector by LSTD.
Hence,
we propose to further transfer warm-up detector as target detector (Fig. \ref{wstdshow}),
which can leverage the reliable object supervision to enhance weakly-supervised detection in the target domain.

\textbf{Object Supervision from Warm-Up Detector}.
We fix warm-up detector as an extractor of object knowledge.
First,
we use the pretrained RPN of warm-up detector as an online proposal generator,
which can produce high-quality proposals for weakly-annotated images in the target domain.
Second,
we feed these proposals into the ROI pooling layer of warm-up detector,
and subsequently generate source detection knowledge $\mathbf{S}^{sdk}$ from the SDK prediction branch.

\textbf{Integrating Warm-Up Supervision into Target Detector}.
We add an extra FC layer as the SDK prediction branch of target detector.
This branch can produce a prediction matrix $\mathbf{P}^{sdk}\in\mathbb{R}^{(C_{s}+1)\times K}$ for object proposals of weakly-annotated images.
By using cross entropy between $\mathbf{S}^{sdk}$ and $\mathbf{P}^{sdk}$,
we integrate warm-up supervision into target detector,
\begin{equation}
\mathcal{L}_{wstd}^{sdk}=\sum\nolimits_{k=1}^{K}\sum\nolimits_{c=1}^{C_{s}+1}\mathbf{S}^{sdk}(c,k)\log{\mathbf{P}^{sdk}(c,k)}.
\label{Eq:TKLoss}
\end{equation}

\textbf{Discussion}.
To further stabilize target detector with weakly-annotated images,
we propose to leverage object supervision from warm-up detector.
Note that,
this supervision refers to source detection knowledge (SDK).
It is transferred from source detector to warm-up detector (by Eq.(\ref{warmloss})),
and then from warm-up detector to target detector (by Eq. (\ref{Eq:TKLoss})).
In other words,
we use warm-up detector as a middle-stage,
instead of directly transferring from source detector to target detector.
Our key insight is that,
warm-up detector is fine-tuned with fully-annotated target images during LSTD,
i.e.,
it has been adaptively adjusted to produce effective target-domain proposals,
and can reliably represent source-domain knowledge of these target proposals.
Via such progressive learning,
one can gradually promote the generalization capacity of object detector in the target domain.

\begin{figure*}[t]
\centering
\includegraphics[width=0.7\textwidth]{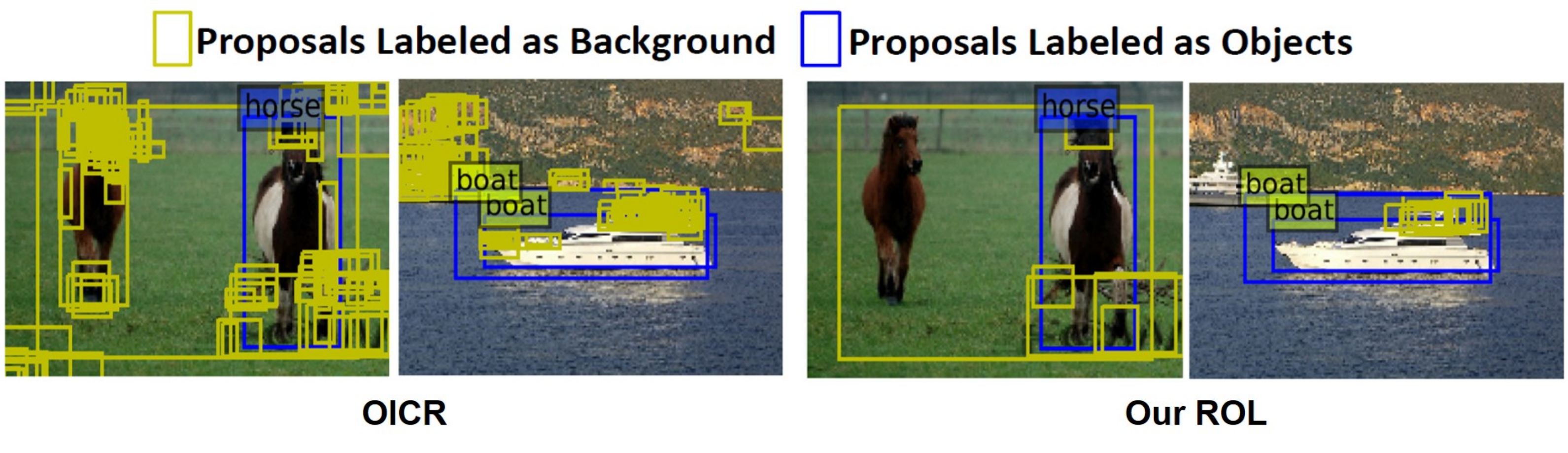}
\caption{Visualization of support proposal mining (OICR \cite{8099809} vs our ROL, both use 128 proposals).
OICR labels many redundant proposals and true objects (e.g., left horse and left boat in the 1st and 2nd images) as background.
Alternatively, our ROL selectively labels background regions around the confident proposals,
which greatly improves the quality of training samples to refine object classifier.
}
\label{ohemshow}
\end{figure*}

\subsection{Recurrent Object Labelling (ROL)}
\label{Recurrent Object Labelling (ROL)}
For learning to detect with weakly-annotated images,
we leverage source domain knowledge (i.e., object supervision of warm-up detector) in the previous section.
Next,
we integrate target domain knowledge for weakly-supervised detection.
Specifically,
we propose a recurrent object labelling (ROL) mechanism,
which can exploit support proposals of weakly-annotated images,
and refine object classifier online.

\textbf{Classifier $i = 1$: Image-Level Supervision}.
After generating proposals of a weakly-annotated image,
we feed them into ROI pooling of target detector (Fig. \ref{wstdshow}).
For each proposal,
conv13 in the target detector can generate a convolutional feature cube.
We subsequently feed these cubes into recurrent object labelling (Fig. \ref{rolvisualize}),
which can identify confident proposals for classifier refinement.
Specifically,
we pass all the convolutional cubes through Classifier $i=1$,
and obtain a prediction matrix $\mathbf{P}^{1}\in\mathbb{R}^{(C_{t}+1)\times K}$.
Each column of $\mathbf{P}^{1}$ refers to the probability vector of a proposal,
and $C_{t}$ is the number of object categories in the target domain.
Note that,
we only have ground truth image label $\mathbf{Y}^{img}\in\mathbb{R}^{C_{t}\times 1}$,
since each image is weakly annotated in the stage of WSTD.
To integrate this supervision into training,
we sum $\mathbf{P}^{1}$ over proposals and obtain an image score vector $\mathbf{P}^{img} \in\mathbb{R}^{(C_{t}+1)\times 1}$,
\begin{equation}
\mathbf{P}^{img}=Sigmoid\{\sum\nolimits_{k=1}^{K}\mathbf{P}^{1}(:,k)\}.
\label{eq:imlep}
\end{equation}
Since there may be multiple objects in one image (i.e., multiple entries of $\mathbf{Y}^{img}$ can be 1),
we apply the multi-label loss for Classifier $i=1$,
\begin{equation}
\mathcal{L}^{1}_{rol}=\sum\nolimits_{c=1}^{C_{t}}\{\mathbf{Y}^{img}_{c}\log {\mathbf{P}^{img}_{c}}+ (1-\mathbf{Y}^{img}_{c})\log{(1-\mathbf{P}^{img}_{c})}\}.
\label{eq:LS1}
\end{equation}
Via this image-level supervision,
we can enhance reliability of proposal score $\mathbf{P}^{1}$,
which will be used for object labelling afterwards.

\textbf{Classifier $i>1$: Support Proposal Mining}.
After obtaining the proposal score $\mathbf{P}^{1}$,
we label object proposals in a recurrent manner.
It is worth mentioning that,
not all the proposals are confident enough to describe objects or background in an image.
Hence,
we design a support proposal mining procedure,
where
we label the highly-confident object and background proposals,
according to the proposal score matrix $\mathbf{P}^{i}$.
In the following,
we illustrate the entire labelling procedure at Classifier $i$,
where we assume that the object category $c$ exists in an image.


\textbf{(1) Labelling Support Object Proposals}.
First,
we find the $c$-th row of the prediction score matrix $\mathbf{P}^{i-1}$,
and then pick out the most confident proposal $j$ which has the highest score,
\begin{equation}
j=\arg\max_{k}\{\mathbf{P}^{i-1}(c,k)\}.
\label{eq:highestS}
\end{equation}
Subsequently,
we label this proposal as `ground truth box' for category $c$,
and use its score $\mathbf{P}^{i-1}(c,j)$ as the pseudo label $\mathbf{S}^{i-1}(c,j)$.
Moreover,
the spatial context is often important for weakly-supervised detection.
Hence,
we further assign category $c$ to those proposals,
which are spatially adjacent to the highest-score proposal $j$ (e.g., IoU$>0.5$).

\textbf{(2) Labelling Support Background Proposals}.
Traditionally,
all the unlabelled proposals are assigned into the background category \cite{8099809}.
This design apparently introduces a large number of wrong annotations,
since the unlabelled proposals, which are far from the highest-score proposal $j$, may also contain objects.
To alleviate it,
we propose to exploit support background proposals from the rest unlabelled ones.
Specifically,
we assign the background category to the ones which have the moderate overlap (e.g., $0.3<$IoU$<0.5$) with the highest-score proposal $j$.
As a result,
we can reduce wrong or redundant annotations which often lead to the unstable learning for weakly-supervised detection.

\textbf{(3) Classification Loss}.
After labelling all the object categories in a weakly-annotated image,
we can get the score matrix $\mathbf{S}^{i}$ as pseudo label.
For the support proposals of category $c$,
we assign the highest score $\mathbf{P}^{i-1}(c,j)$ into the corresponding entries of $\mathbf{S}^{i}$.
This design can effectively associate the object category with its confident proposals,
which stabilizes the training procedure of weakly-supervised detection \cite{8099809}.
For the unlabelled proposals,
the corresponding columns are zero vectors in $\mathbf{S}^{i}$.
Finally,
we obtain the prediction score matrix $\mathbf{P}^{i}$ from Classifier $i$,
and compute the cross entropy loss between $\mathbf{P}^{i}$ and $\mathbf{S}^{i}$ for training,
\begin{equation}
\mathcal{L}^{i}_{rol}=\sum\nolimits_{k=1}^{K}\sum\nolimits_{c=1}^{(C_{t}+1)}\mathbf{S}^{i}(c,k)\log{\mathbf{P}^{i}(c,k)}.
\label{eq:LSi}
\end{equation}
This procedure is done in a recurrent manner,
and the total loss of our ROL module is
\begin{equation}
\mathcal{L}^{rol}_{wstd}=\sum\nolimits_{i}\mathcal{L}^{i}_{rol}.
\label{eq:Ltotal}
\end{equation}

\begin{table*}[t]
\centering
\caption{Data Settings of LSTD.
The object categories for source and target are carefully selected to be non-overlapped,
in order to evaluate if LSTD can detect unseen object categories from few training shots in the target domain.}
\begin{tabular}{l|l|l}
  \hline
  LSTD & Source (large-scale, fully-annotated) & Target (low-shot, fully-annotated) \\
  \hline
  Task 1 & COCO (standard 80 classes, 118,287 training images) &  ImageNet2015 (chosen 50 classes) \\
  Task 2& COCO (chosen 60 classes, 98,459 training images)   &  VOC2007 (standard 20 classes) \\
  Task 3& ImageNet2015 (chosen 181 classes, 151,603 training images) & VOC2010 (standard 20 classes) \\
  \hline
\end{tabular}
\label{DataDescription}
\end{table*}

\begin{table}[t]
\centering
\caption{Basic deep structure of LSTD.
We compare LSTD with its closely-related SSD \cite{Liueccv2016} and Faster RCNN \cite{Renpami2016} in both source and target domains.
For fairness,
we choose task 1 to show the effectiveness.
The main reason is that,
the source data in task 1 is the standard COCO detection set,
where SSD and Faster RCNN are well-trained with the state-of-the-art performance.
Hence,
we use the published SSD \cite{Liueccv2016} and Faster RCNN \cite{Renpami2016} in this experiment.
The mAP results show that,
LSTD is a more effective deep architecture for low-shot detection in the target domain.
Furthermore,
the structure of LSTD itself is robust,
with regards to different convolutional layers for ROI pooling.}
\resizebox{0.47\textwidth}{!}{
\begin{tabular}{l|cc}
  \hline
  Deep Models (mAP) & large-scale source & low-shot target\\
  \hline
  Faster RCNN \cite{Renpami2016} & 21.9  & 12.2  \\
  SSD \cite{Liueccv2016} & 25.1 & 10.1 \\
  \hline
  Our LSTD$_{conv5_{-}3}$ & 24.7  & 15.9  \\
  Our LSTD$_{conv7}$ & \textbf{25.2}  & \textbf{16.5}  \\
  \hline
\end{tabular}
}
\label{LSTDSourceTarget}
\end{table}

\begin{table}[t]
\centering
\caption{Regularized transfer learning for LSTD.
$FT$: standard fine-tuning.
$SDK$: source detection knowledge (SDK) regularization.
$BD$: background depression (BD) regularization.
The mAP results show that,
both regularization terms can significantly help the fine-tuning procedure of LSTD,
when the training set is scarce in the target domain.}
\resizebox{0.47\textwidth}{!}{
\begin{tabular}{l|ccccc}
\hline
Shots for Task 1 (mAP) & 1 & 2 & 5 & 10 & 30  \\
\hline
LSTD$_{FT}$ &  16.5 & 21.9 & 34.3 & 41.5 & 52.6 \\
LSTD$_{FT + SDK}$ & 18.1 & 25.0 & 35.6 & 43.3 & 55.0 \\
LSTD$_{FT + SDK + BD}$ & \textbf{19.2} & \textbf{25.8} & \textbf{37.4} & \textbf{44.3} & \textbf{55.8} \\
\hline
Shots for Task 2 (mAP) & 1 & 2 & 5 & 10 & 30  \\
\hline
LSTD$_{FT}$& 27.1 & 46.1 & 57.9 & 63.2 & 67.2 \\
LSTD$_{FT + SDK}$ & 31.8 & 50.7 & 60.4 & 65.1 & 69.0 \\
LSTD$_{FT + SDK + BD}$ & \textbf{34.0} & \textbf{51.9} & \textbf{60.9} & \textbf{65.5} & \textbf{69.7} \\
\hline
Shots for Task 3 (mAP) & 1 & 2 & 5 & 10 & 30  \\
\hline
LSTD$_{FT}$& 29.3 & 37.2  & 48.1 & 52.1 & 56.4 \\
LSTD$_{FT + SDK}$ & 32.7 & 40.8 & 49.7 & 54.1 & 57.9 \\
LSTD$_{FT + SDK + BD}$ & \textbf{33.6} & \textbf{42.5} & \textbf{50.9} & \textbf{54.5} & \textbf{58.3} \\
\hline
\end{tabular}
}
\label{LSTDTransfer}
\end{table}

\begin{table}[t]
\centering
\caption{Background Depression (BD) regularization.
We perform BD regularization on different convolutional layers when fine-tuning.
The mAP results show that BD is robust to different convolutional layers.}
\begin{tabular}{l|ccccc}
\hline
Tasks (mAP) & $BD_{conv5_{-}3}$ & $BD_{conv7}$  \\
\hline
Task1 & \textbf{19.2}  & 18.9 \\
Task2 & 34.0  & \textbf{34.5}  \\
Task3 & \textbf{33.6}  & 33.4 \\
\hline
\end{tabular}
\label{BDdiffconv}
\end{table}

\textbf{Discussion}.
First,
we clarify the main difference between Online Instance Classifier Refinement (OICR) \cite{8099809} and our ROL.
Specifically,
OICR is a label propagation and refinement technique for weakly-supervised detection.
Similar to our ROL,
it labels support object proposals via IOU.
However,
OICR assigns the background category without selection,
while our ROL takes support proposal mining into account.
As shown in Fig. \ref{ohemshow},
OICR labels many redundant proposals as background.
What is worse,
it tends to label true objects (e.g., left horse in the 1st image) mistakenly as background,
and these noisy annotations can deteriorate the refining procedure of object classifier.
On the contrary,
our ROL carefully labels the contextual background regions around the confident proposals,
which can reduce the labelling redundancy and improve the quality of training samples.
Finally,
we would like to emphasize that,
our WSTD is used to imitate the generalization process of human learning,
i.e.,
humans generalize the warm-up stage by exploiting objects from wild images without full annotations.
To achieve it,
WSTD investigates reliable object supervision from warm-up detector of LSTD.
More importantly,
it uses ROL for learning to detect objects from weakly-annotated images.
In this case,
WSTD can further generalize the detection performance in the target domain.
Next,
we evaluate progressive object transfer detection (POTD),
by extensive experiments on a number of challenging datasets.

\section{Experiments}

In this section,
we evaluate the performance of the proposed POTD on a number of challenging data settings.
First,
since POTD consists of two transfer detection stages (i.e., LSTD and WSTD),
we deeply investigate them from different experimental aspects.
Then,
we compare our POTD with the state-of-the-art approaches,
and show our contributions in practice.

\subsection{LSTD}

\textbf{Data Settings}.
Since LSTD is a regularized transfer learning framework for low-shot detection,
we adopt a number of detection benchmarks,
i.e.,
COCO \cite{coco},
ImageNet2015 \cite{imagenet},
VOC2007 and VOC2010 \cite{voc},
respectively as source and target of three transfer tasks (Table \ref{DataDescription}).
The training set is large-scale in the source domain of each task,
while
it is low-shot in the target domain (1/2/5/10/30 training images for each target-object class).
The fully-annotated training shots are randomly selected in our experiments.
To reproduce the results,
we provide the data splits for all the tasks\footnote{$https://github.com/Cassie94/LSTD/tree/master/data\_split$}.
Furthermore,
the object categories for source and target are carefully selected to be non-overlapped,
in order to evaluate if LSTD can detect unseen object categories from few shots in the target domain.
Finally,
we use the standard PASCAL VOC detection rule on the test sets to report mean average precision (mAP) with 0.5 intersection-over-union (IOU).


\textbf{Implementation Details}.
Unless stated otherwise,
we perform LSTD as follows.
\textbf{First},
the basic deep architecture of LSTD is built upon VGG16 \cite{simonyan2014very},
similar to SSD and Faster RCNN.
For bounding box regression,
we use the same structure in the standard SSD.
For object classification,
we apply the ROI pooling layer on conv7,
and add two convolutional layers (conv12: $3\times3\times256$, conv13: $3\times3\times256$ for task 1/2/3) before object classifier.
\textbf{Second},
we select 100/100/64 proposals (task 1/2/3) for training source detector,
while
we select 64/64/64 proposals for warm-up detector.
The loss coefficients of both BD and SDK are 0.5.
\textbf{Finally},
the optimization strategy for both source and target is Adam \cite{Kingma2015},
where
the initial learning rate is 0.0002 (with 0.1 decay),
the momentum/momentum2 is 0.9/0.99,
and the weight decay is 0.0001.
All our experiments are implemented on Caffe \cite{jia2014caffe}.
In the following,
we evaluate the key designs of LSTD.
To be fair,
when we explore different settings of one design,
others are with the basic setting above.

\textbf{Basic Deep Structure of LSTD}.
We first evaluate the basic deep structure of LSTD respectively in the source and target domains,
where we compare it with the closely-related SSD \cite{Liueccv2016} and Faster RCNN \cite{Renpami2016}.
For fairness,
we choose task 1 to show the effectiveness.
The main reason is that,
the source data in task 1 is the standard COCO detection set,
where SSD and Faster RCNN are well-trained with the state-of-the-art performance.
Hence,
we use the published SSD \cite{Liueccv2016} and Faster RCNN \cite{Renpami2016} in this experiment,
where
the size of input images for SSD and our LSTD is $300\times300$,
and Faster RCNN follows the settings in the original paper.
In Table \ref{LSTDSourceTarget},
we report mAP on the test sets of both source and target domains in task 1.
One can see that,
our LSTD achieves a competitive mAP in the source domain.
It illustrates that LSTD can be a state-of-art deep detector for large-scale training sets.
More importantly,
our LSTD outperforms both SSD and Faster RCNN significantly for low-shot detection in the target domain (one training image per target category),
where
all approaches are simply fine-tuned from their pre-trained models in the source domain.
It shows that,
LSTD yields a more effective deep architecture for low-shot detection,
by leveraging the core designs of SSD and Faster RCNN.
Finally,
we investigate the structure robustness in LSTD itself.
As the bounding box regression follows the standard SSD,
we explore the object classifier in which we choose different convolutional layers (conv$5_{-}3$ or conv$7$) for ROI pooling.
The results are comparable in Table \ref{LSTDSourceTarget},
showing the architecture robustness of LSTD.
For consistency,
we use conv$7$ for ROI pooling in all our experiments.

\textbf{Regularized Transfer Learning of LSTD}.
We mainly evaluate if the proposed regularization can enhance transfer learning for LSTD,
in order to boost low-shot detection.
As shown in Table \ref{LSTDTransfer},
both SDK and BD can significantly improve the baseline (i.e., direct fine-tuning),
especially when the training set is scarce in the target domain (such as one-shot).
Additionally,
we show the architecture robustness of BD regularization in Table \ref{BDdiffconv}.
Specifically,
we perform BD regularization on different convolutional layers.
One can see that BD is generally robust to different convolutional layers.
Hence,
we apply BD on conv$5_{-}3$ in our experiments for consistency.



\begin{table}[t]
\centering
\caption{WSTD vs. LSTD.
After applying LSTD with few fully-annotated images (i.e., 1/2/5/10/30 shots per class),
we continue to perform WSTD with weakly-annotated images (i.e., 0/$\sim$250-shot(VOC07 Training Data)/$\sim$800-shot(VOC07+12 Training Data) per class).
More explanations can be found in the text.}
\resizebox{0.48\textwidth}{!}{
\begin{tabular}{c|l|c|l|l}
\hline\hline
LSTD (Full Annotation)   & \multicolumn{1}{c|}{WSTD (Weak Annotation)} & \multicolumn{3}{c}{mAP}  \\ \hline\hline
\multirow{3}{*}{1-shot}  & 0-shot                                      & \multicolumn{3}{c}{34.0} \\
                         & $\sim$250-shot (VOC07)                      & \multicolumn{3}{c}{62.6} \\
                         & $\sim$800-shot (VOC07+12)                   & \multicolumn{3}{c}{62.8} \\ \hline\hline
\multirow{3}{*}{2-shot}  & 0-shot                                      & \multicolumn{3}{c}{51.9} \\
                         & $\sim$250-shot (VOC07)                      & \multicolumn{3}{c}{65.6} \\
                         & $\sim$800-shot (VOC07+12)                   & \multicolumn{3}{c}{66.0} \\ \hline\hline
\multirow{3}{*}{5-shot}  & 0-shot                                      & \multicolumn{3}{c}{60.9} \\
                         & $\sim$250-shot (VOC07)                      & \multicolumn{3}{c}{69.1} \\
                         & $\sim$800-shot (VOC07+12)                   & \multicolumn{3}{c}{69.6} \\ \hline\hline
\multirow{3}{*}{10-shot} & 0-shot                                      & \multicolumn{3}{c}{65.5} \\
                         & $\sim$250-shot (VOC07)                      & \multicolumn{3}{c}{69.5} \\
                         & $\sim$800-shot (VOC07+12)                   & \multicolumn{3}{c}{70.1} \\ \hline\hline
\multirow{3}{*}{30-shot} & 0-shot                                      & \multicolumn{3}{c}{69.7} \\
                         & $\sim$250-shot (VOC07)                      & \multicolumn{3}{c}{70.5} \\
                         & $\sim$800-shot (VOC07+12)                   & \multicolumn{3}{c}{71.3} \\ \hline\hline
\end{tabular}
}
\label{shotno}
\end{table}

\begin{table}[t]
\centering
\caption{ROL vs. OICR \cite{8099809}.
OICR is Online Instance Classifier Refinement,
which addresses weakly-supervised detection via label propagation and refinement \cite{8099809}.
It can be used with any detection backbone,
just like our ROL.
Hence,
We perform WSTD respectively with OICR \cite{8099809} and our ROL,
and evaluate them on all the classifiers in the recurrent steps.}
\resizebox{0.45\textwidth}{!}{
\begin{tabular}{l|ccc}
\hline
mAP of WSTD & Classifier 1  & Classifier 2  & Classifier 3 \\
\hline
OICR  & 52.2  & 55.4 & 55.5 \\
Our ROL & \textbf{53.8}  & \textbf{60.7}  & \textbf{61.3} \\
\hline
\end{tabular}
}
\label{SupportMining}
\end{table}

\begin{figure}[t]
\centering
\includegraphics[width=0.35\textwidth]{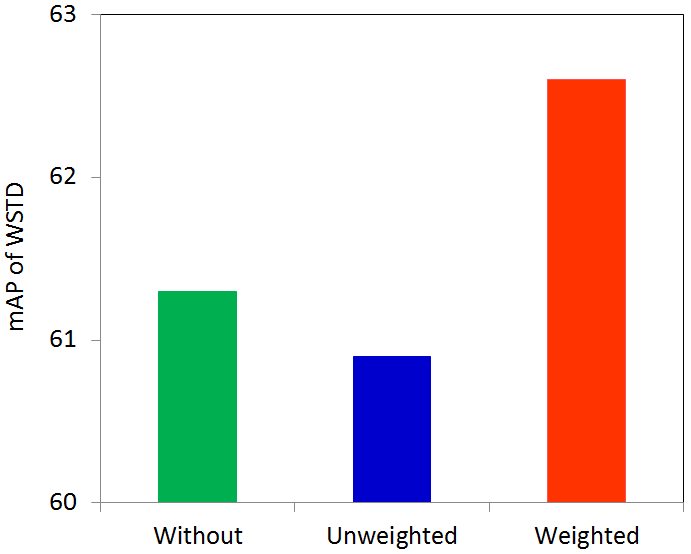}
\caption{Object Supervision of Warm-Up Detector for WSTD.
\emph{SDK (without)} denotes that we ignore SDK in warm-up detector when performing WSTD.
\emph{SDK (unweighted)} denotes that we use the SDK loss in Eq. (\ref{Eq:TKLoss}) when performing WSTD.
\emph{SDK (weighted)} denotes that we add an extra weight in SDK loss,
i.e.,
$\alpha(c,k)\mathbf{S}^{sdk}(c,k)$ is used as object supervision in  Eq. (\ref{Eq:TKLoss}).
In our experiment,
$\alpha(c,k)$ is set as $\mathbf{S}^{sdk}(c,k)$,
for further enhancing the importance of SDK.}
\label{wstdsdk}
\end{figure}

\begin{figure*}[t]
\centering
\includegraphics[width=0.85\textwidth]{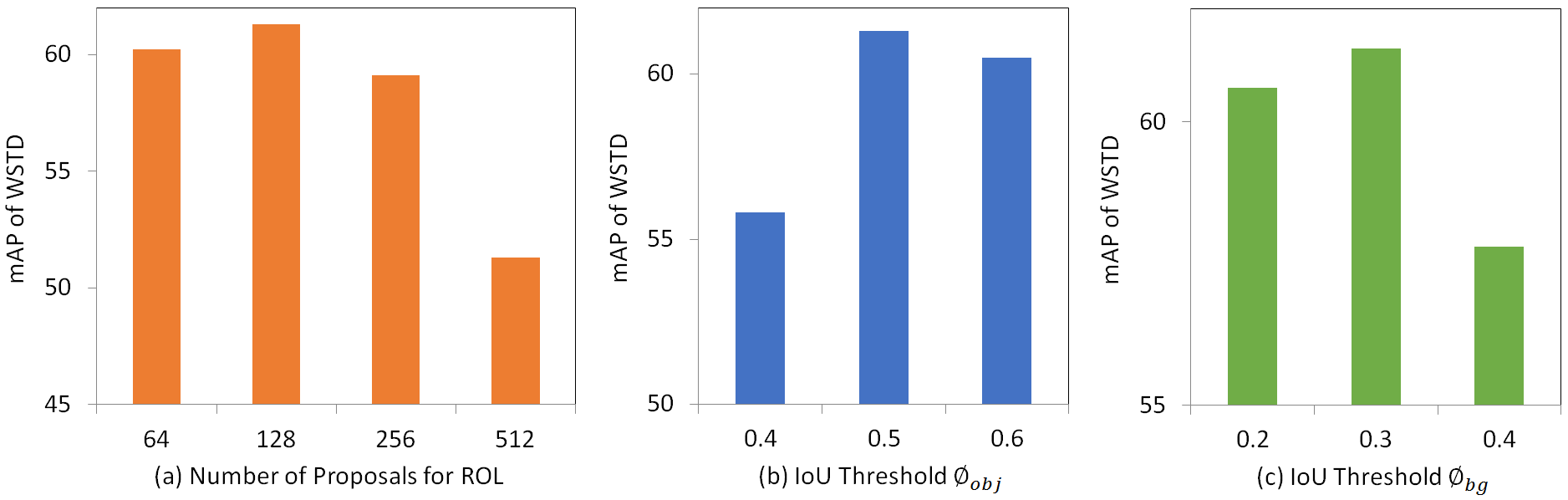}
\caption{Recurrent Object Labelling (ROL) for WSTD.
First,
mAP of WSTD first increases and then decreases,
when we increase the number of proposals for ROL.
It illustrates that,
the number of proposals is required to be sufficient for a good detection performance.
But too many proposals may introduce noisy annotations to deteriorate ROL.
Second,
mAP of WSTD first increases and then decreases,
as $\phi_{obj}$ or $\phi_{bg}$ increases.
It shows that,
the threshold should not be too loose or tight for support proposal mining.
Hence,
we choose 128 proposals, $\phi_{obj}=0.5$ and $\phi_{bg}=0.3$ in the rest experiments,
due to the outstanding performance of this setting.}
\label{wstdrol}
\end{figure*}

\begin{table*}[t]
\centering
\caption{WSTD (with ROL or OICR) on each category of VOC2007. $2/5shot$ denotes that only 2/5 training images are fully annotated with bounding boxes, all other training images are weakly-annotated with image labels.}
\resizebox{\textwidth}{!}{
\begin{tabular}{l| *{20}{c} |c}
 \hline
Methods &aero&bike&bird &boat&bottle&bus&car&cat&chair&cow&
	table&dog&horse&mbike&person&plant&sheep&sofa&train&tv &mAP\\
\hline
OICR$_{(2shot)}$  &64.6& 69.6& 50.3& 49.7& 30.4& 77.5& 68.0& 74.1& 39.5& 58.9& 53.4& 70.5& 77.3& 67.9& 50.0& 31.1& 52.3& 73.7& 76.3& 62.1& 59.9\\
OICR$_{(5shot)}$  &70.4& 72.0& 57.7& 56.6& 33.8& 75.6& 69.7& 80.0& \textbf{42.8}& 59.9& \textbf{67.4} & 75.6& 76.4& 68.5& 53.5& 32.8& 52.8& 71.5& 77.5& 65.8& 63.0 \\
\hline
Our ROL$_{(2shot)}$& 68.9  & 73.0 & 56.0 & 54.3 & 35.5 & 81.9 & 79.2 & 75.9 & 41.1 & 72.8 & 56.9 & 74.9 & \textbf{83.0} & 75.3 & 64.0 & 35.7 & 64.6 & \textbf{74.5} & 77.5 & 67.9  & 65.6 \\
Our ROL$_{(5shot)}$ &\textbf{75.1} &\textbf{76.3 }&\textbf{66.2} &\textbf{58.6} &\textbf{42.6} &\textbf{80.1} &\textbf{80.0} &\textbf{82.6} &39.7 &\textbf{76.1} &66.1 &\textbf{80.5} &82.4 &\textbf{78.9} &\textbf{68.6} &\textbf{38.6} &\textbf{72.0} &70.2 &\textbf{78.4} &\textbf{69.6} &\textbf{69.1} \\
\hline
\end{tabular}
}
\label{aavoc2007}

\vspace{0.3cm}

\caption{WSTD (with ROL or OICR) on each category of VOC2010. $2/5shot$ denotes that only 2/5 training images are fully annotated with bounding boxes, all other training images are weakly-annotated with image labels.}
\resizebox{\textwidth}{!}{
\begin{tabular}{l| *{20}{c} |c}
 \hline
Methods &aero&bike&bird &boat&bottle&bus&car&cat&chair&cow&
	table&dog&horse&mbike&person&plant&sheep&sofa&train&tv &mAP\\
\hline
OICR$_{(2shot)}$  &73.3& 64.1& 50.8& 44.1& 31.9& 64.5& 55.8& 72.3& 34.3& 52.8& 31.9& 73.6& 71.1& 69.0& 52.2& 29.3& 45.0& 59.2& 72.9& 53.0& 55.1\\
OICR$_{(5shot)}$ &75.3& 66.1& 56.0& 45.0& 35.0& 62.7& 56.1& 76.7& 34.2& 53.8& \textbf{48.8}& 75.6& 69.9& 70.5& 53.6& 31.3& 47.9& 55.6& 73.4& 55.0& 57.1 \\
\hline
Our ROL$_{(2shot)}$& 77.7& 69.2& 57.0& \textbf{48.3}& 33.6& 73.9& 67.5& 77.1& \textbf{35.1}& 67.7& 34.6& 78.5& 78.2& 73.9& 65.9& 34.1& 62.4& \textbf{59.5}& 76.8& 55.4& 61.3 \\
Our ROL$_{(5shot)}$ &\textbf{81.3} & \textbf{72.9}& \textbf{62.6} & 47.0& \textbf{37.9}& \textbf{74.1}& \textbf{69.2}& \textbf{80.0}& 28.9& \textbf{68.6}& 48.0& \textbf{81.1}& \textbf{78.5}& \textbf{78.9}& \textbf{70.7}& \textbf{37.0}& \textbf{68.9}& 55.8& \textbf{77.0}& \textbf{56.7}& \textbf{63.8} \\
\hline
\end{tabular}
}
\label{aavoc2010}

\vspace{0.3cm}

\caption{WSTD (with ROL or OICR) on each category of VOC2012. $2/5shot$ denotes that only 2/5 training images are fully annotated with bounding boxes, all other training images are weakly-annotated with image labels.}
\resizebox{\textwidth}{!}{
\begin{tabular}{l| *{20}{c} |c}
 \hline
Methods &aero&bike&bird &boat&bottle&bus&car&cat&chair&cow&
	table&dog&horse&mbike&person&plant&sheep&sofa&train&tv &mAP\\
\hline
OICR$_{(2shot)}$  &72.3& 63.8& 50.0& 41.3& 29.3& 64.8& 55.0& 71.9& 33.4& 51.8& 31.8& 73.5& 70.4& 68.0& 51.0& 28.9& 43.5& 59.8& 72.4& 53.7& 54.3 \\
OICR$_{(5shot)}$  &73.7& 66.5& 55.0& 41.3& 32.1& 63.1& 55.2& 76.1& 33.2& 52.7& \textbf{48.9}& 75.3& 69.6& 69.1& 52.3& 30.0& 46.9& 57.1& 73.5& 55.8& 56.4 \\
\hline
Our ROL$_{(2shot)}$& 77.1& 68.8& 55.8& \textbf{45.0}& 30.7& 74.0& 67.8& 76.8& \textbf{34.8}& 67.4& 34.8& 78.3& 77.7& 73.5& 65.3& 33.3& 60.9& \textbf{60.2}& 75.9& 56.2& 60.7 \\
Our ROL$_{(5shot)}$ &\textbf{80.2}& \textbf{73.4}& \textbf{61.6}& 43.0& \textbf{35.2}& \textbf{74.5}& \textbf{69.4}& \textbf{79.6}& 28.8& \textbf{68.1}& 48.3& \textbf{80.8}& \textbf{78.5}& \textbf{78.0}& \textbf{70.0}& \textbf{35.8}& \textbf{67.8}& 56.7& \textbf{76.6}&\textbf{57.8}& \textbf{63.2} \\
\hline
\end{tabular}
}
\label{aavoc2012}

\end{table*}

\subsection{WSTD}

\textbf{Data Settings}.
We mainly evaluate WSTD on task 2,
since it is the most representative setting among the three tasks in the previous section.
First,
one image contains multiple objects in COCO and VOC2007,
while
one image contains one object in ImageNet2015.
Hence,
task 2 is a more realistic data setting.
Second,
the target domain refers to VOC2007 with the standard 20 categories,
which is more convenient to compare with other approaches.
Furthermore,
we extend the target domain of task 2 as VOC2010 and VOC2012 later on,
in order to further show the effectiveness of WSTD.

\textbf{Implementation Details}.
First,
we start WSTD from warm-up detector,
which is trained with 1 fully-annotated image per category of VOC2007 in the stage of LSTD.
Then,
we utilize WSTD to train target detector,
where
the training images are weakly annotated.
Second,
we choose 128 proposals for each image in WSTD,
after non-maximum-suppression (NMS) of 1500 proposals at 0.75.
Moreover,
we simultaneously enlarge the loss coefficients in WSTD (i.e., 50 for ROL and 150 for SDK),
in order to speed up convergence.
Finally,
the optimization strategy is Adam \cite{Kingma2015},
where the initial learning rate is 0.00001 (with 0.1 decay),
and the weight decay is 0.0001.
All other settings are the same as LSTD.

\textbf{WSTD vs. LSTD}.
After applying LSTD with few fully-annotated images,
we continue to perform WSTD with weakly-annotated images.
The result is shown in Table \ref{shotno}.
\textbf{First},
when the number of fully-annotated training images in LSTD is small,
one can perform WSTD to improve the detection mAP with weakly-annotated images.
For example,
the mAP is 34.0 when we fully annotate 1 shot per class in LSTD.
It becomes as 62.6 after we use 250 weakly-annotated images per class in WSTD,
and tends to stabilize around 62.8 after we use 800 weakly-annotated images per class in WSTD.
It illustrates that,
WSTD can significantly boost low-shot detection via weakly-annotated images,
and tends to be saturated when the number of weakly-annotated images increases.
\textbf{Second},
when the number of fully-annotated training images in LSTD is getting larger,
the efforts of weakly-annotated images in WSTD is getting smaller.
For example,
the detection mAP is 69.7 when we fully annotate 30 shots per class in LSTD.
After we use around 250 weakly-annotated images per class in WSTD,
the mAP slightly increases to 70.5.
It illustrates that,
fully-annotated images take more effort for detection in the target domain.
\textbf{Third},
there exists the tradeoff between LSTD and WSTD.
For example,
the mAP is 70.1,
when we use 10 fully-annotated images (per class) in LSTD and around 800 weakly-annotated images (per class) in WSTD.
Alternatively,
we can obtain the similar mAP (70.5),
when we use 30 fully-annotated images (per class) in LSTD and around 250 weakly-annotated images (per class) in WSTD.
Since weakly-annotated images can be straightforwardly obtained from internet,
while
fully-annotated images have to be obtained via the exhausted labelling procedure.
Hence,
it is preferable choice to use few fully-annotated images in LSTD and most weakly-annotated images in WSTD.


\textbf{Object Supervision from Warm-Up Detector}.
We evaluate whether object supervision of warm-up detector is effective for WSTD.
Specifically,
we consider three cases in the following.
\emph{SDK (without)} denotes that we ignore SDK in warm-up detector when performing WSTD.
\emph{SDK (unweighted)} denotes that we use the SDK loss in Eq. (\ref{Eq:TKLoss}) when performing WSTD.
\emph{SDK (weighted)} denotes that we add an extra weight in SDK loss,
i.e.,
$\alpha(c,k)\mathbf{S}^{sdk}(c,k)$ is used as object supervision in  Eq. (\ref{Eq:TKLoss}).
In our experiment,
$\alpha(c,k)$ is set to be $\mathbf{S}^{sdk}(c,k)$,
in order to further enhance the importance of SDK.
The result is shown in Fig. \ref{wstdsdk}.
One can see that,
\emph{SDK (unweighted)} is comparable to \emph{SDK (without)}.
It illustrates that SDK needs to be further exploited in the object-level.
Furthermore,
\emph{SDK (weighted)} achieves the best among these settings,
showing that $\alpha(c,k)$ can further take the importance of SDK into account.

\textbf{Recurrent Object Labelling (ROL)}.
We mainly evaluate ROL,
w.r.t.,
number of proposals,
IoU threshold,
and ROL vs. OICR \cite{8099809}.
\textbf{(1) Number of Proposals}.
As shown in Fig. \ref{wstdrol} (a),
mAP of WSTD first increases and then decreases,
when we increase the number of proposals for ROL.
It illustrates that,
the number of proposals is required to be sufficient for a good detection performance.
But too many proposals may introduce noisy annotations to deteriorate ROL.
Hence,
we choose 128 proposals in the rest experiments,
due to its outstanding performance.
\textbf{(2) IoU Threshold}.
In ROL,
we use IoU thresholds to selectively mine support proposals as objects (i.e., IoU $>\phi_{obj}$) or background (i.e., $\phi_{bg}<$ IoU $<\phi_{obj}$).
Hence,
we evaluate the influence of IoU thresholds in Fig. \ref{wstdrol} (b)-(c),
where
we change $\phi_{obj}$ for proposals labelled as objects ($\phi_{bg}=0.3$ in this case),
and change $\phi_{bg}$ for proposals labelled as background ($\phi_{obj}=0.5$ in this case).
As $\phi_{obj}$ or $\phi_{bg}$ increases,
mAP of WSTD first increases and then decreases.
It shows that,
the threshold should not be too loose or tight for support proposal mining.
Hence,
we set $\phi_{obj}=0.5$ and $\phi_{bg}=0.3$ in the rest experiments.

\textbf{ROL vs. OICR \cite{8099809}}.
OICR is Online Instance Classifier Refinement,
which addresses weakly-supervised detection via label propagation and refinement \cite{8099809}.
It can be used with any detection backbone,
just like our ROL.
To further show the effectiveness of ROL,
we respectively perform WSTD with OICR \cite{8099809} and our proposed ROL (both utilize 128 proposals).
In Table \ref{SupportMining},
our ROL significantly outperforms OICR for all the classifiers in the recurrent steps.
This is mainly because that,
our ROL selects support proposals attentively,
which can largely reduce redundant proposals and noisy annotations.
Furthermore,
we evaluate ROL vs. OICR \cite{8099809} on each category of VOC2007/2010/2012.
For most object categories in Table \ref{aavoc2007}, \ref{aavoc2010} and \ref{aavoc2012},
our ROL outperforms OICR with a large margin,
showing the effectiveness of ROL.

\begin{table}[t]
\centering
\caption{The mAP Comparison with The-State-of-The-Art.
Note that,
we would like to summarize our contribution as a complete framework,
when comparing with the state-of-the-art approaches.
Hence,
we use the notation of POTD,
instead of WSTD in all the tables before.
One can see that POTD$_{(2shot)}$ is competitive to CBL$_{(2shot)}$,
but it uses much less weakly-annotated data,
i.e.,
POTD vs. CBL: 250 weakly-annotated images vs. 20,000 weakly-annotated videos per category.
Furthermore,
POTD$_{(all)}$ significantly outperforms CBL$_{(all)}$,
and achieves the comparable results to the fully-supervised detectors.}
\resizebox{0.47\textwidth}{!}{
\begin{tabular}{l|c c c }
  \hline
  \hline
   Weakly-Supervised (mAP) & VOC07  &  VOC10 &  VOC12 \\
  \hline
  \hline
   WSDDN \cite{bilen2016weaklyddn}& 34.8 & 36.2 & - \\
	STL  \cite{8099940}& 41.7 & - &  38.3\\
	WCCN \cite{diba2016weakly}& 42.8 & 39.5 &37.9 \\
	PDA  \cite{li2016weakly}& 39.5 &30.7 &39.1 \\
    OICR \cite{8099809}&41.2& - &37.9  \\
	SGW  \cite{Lai2017SaliencyGE}& \textbf{43.5} & -  &\textbf{39.6} \\	
  \hline
  \hline
  Semi-Supervised (mAP)& VOC07  &  VOC10 &  VOC12 \\
  \hline
  \hline
   MSPLD$_{(2shot)}$\cite{dong2017few}&37.4& -& -\\
   MSPLD$_{(3shot)}$\cite{dong2017few}&44.8& -&  -\\
   MSPLD$_{(4shot)}$\cite{dong2017few}&\textbf{47.5}& -&  -\\
  \hline
  \hline
  Fully-Supervised (mAP)& VOC07  &  VOC10 &  VOC12 \\
  \hline
  \hline
  Faster RCNN \cite{Renpami2016} &\textbf{69.9}& - &70.4 (07+12) \\
  SSD \cite{Liueccv2016} &68.0 & -&\textbf{72.4}  (07+12)\\
  \hline
  \hline
  Low-Shot Transfer (mAP)& VOC07  &  VOC10 &  VOC12 \\
  \hline
  \hline
  CBL$_{(2shot)}$\cite{Liang2015TowardsCB} & 67.1 & 63.8  & 63.2\\
  CBL$_{(all)}$\cite{Liang2015TowardsCB} & 68.9 & - & 64.6\\
  Our POTD$_{(1shot)}$&62.6 & 58.6  &   58.1\\
  Our POTD$_{(2shot)}$&65.6 &61.3 &60.7\\
  Our POTD$_{(5shot)}$&69.1 &63.8& 63.2\\
  Our POTD$_{(all)}$&\textbf{76.1}&  \textbf{71.8} &  \textbf{71.3}\\
  \hline
  \hline
\end{tabular}
}
\label{stateoftheart}
\end{table}

\subsection{Comparison with The-State-of-The-Art}

In this section,
we compare our POTD with a number of the state-of-the-art approaches in Table \ref{stateoftheart}.
First,
the performance of weakly-supervised / semi-supervised detectors is far from competitive to fully-supervised detectors,
due to the lack of object-level supervision.
Alternatively,
low-shot transfer detectors can achieve the competitive results,
even though few images are fully-annotated in the target domain.
The main reason is that these transfer detectors can mimic the human learning,
which leverages source-domain knowledge as object prior.
Second,
we compare POTD with the recent computational baby learning (CBL) \cite{Liang2015TowardsCB}.
One can see that,
POTD$_{(2shot)}$ is competitive to CBL$_{(2shot)}$,
but it uses much less weakly-annotated data,
i.e.,
POTD uses 250 weakly-annotated images per category while CBL uses 20,000 weakly-annotated videos per category.
Finally,
the subscript $all$ means that all training images are fully-annotated.
In this case,
POTD is simply reduced as LSTD without using weakly-labeled images.
As expected,
it significantly outperforms CBL$_{(all)}$,
and achieves the comparable results to the fully-supervised detectors.
The main reason is that,
our POTD leverages the key insight of transfer detection,
which can inherit rich source knowledge to boost detection accuracy in the target domain.


\subsection{Visualization}

\begin{figure*}[t]
\centering
\subfigure[Detection Results of LSTD$_{(1shot)}$]{\includegraphics[width=0.98\textwidth]{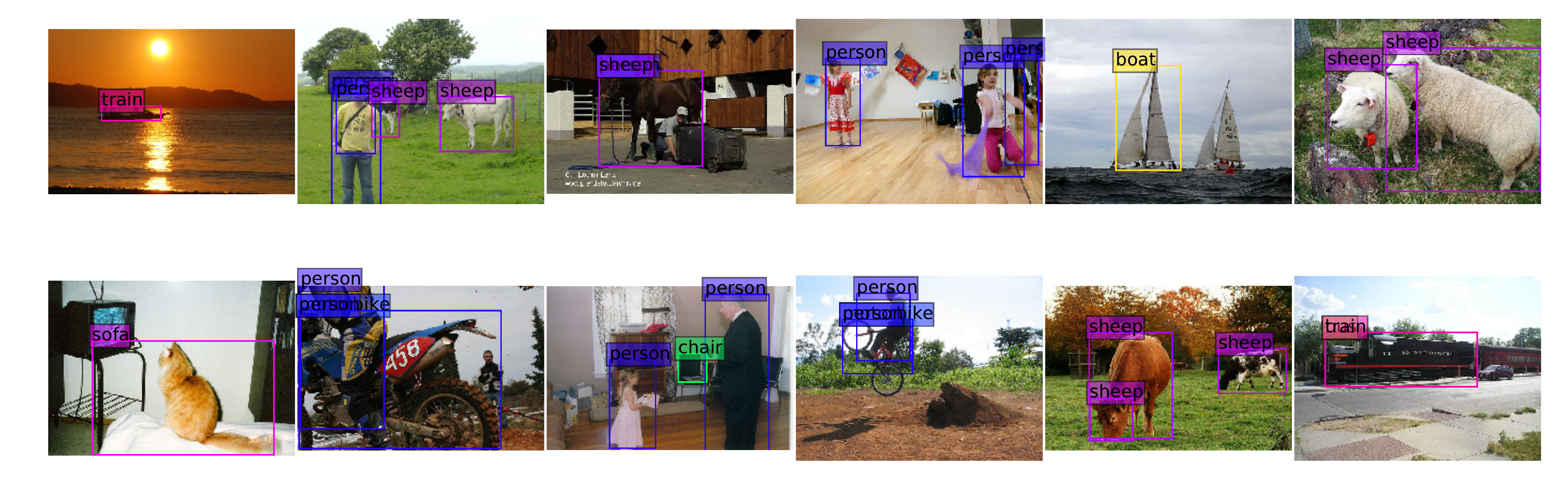}\label{alpha}}
\subfigure[Detection Results of WSTD$_{(1shot)}$]{\includegraphics[width=0.98\textwidth]{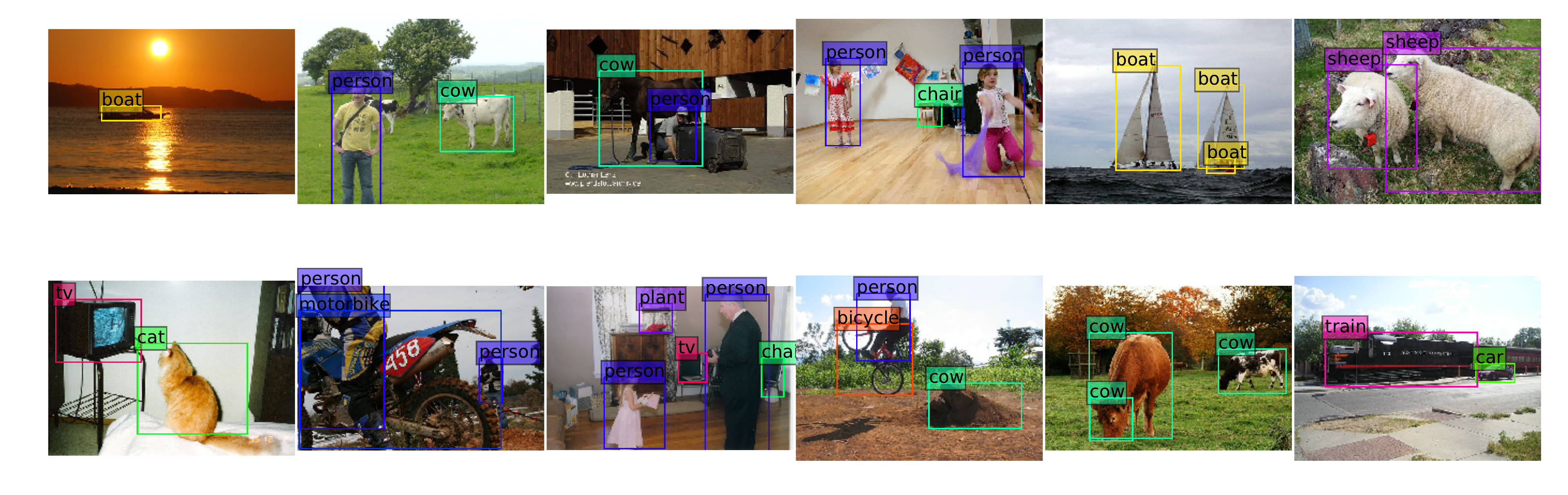}\label{beta}}
\caption{Detection Results of POTD (including two transfer detection stages, i.e., LSTD and WSTD). First, LSTD$_{(1shot)}$ can achieve a reasonable detection performance with only 1 shot fully-annotated images per category. Furthermore, WSTD$_{(1shot)}$ can further generalize the performance of LSTD$_{(1shot)}$ via weakly-annotated images. Hence, our POTD is a preferable deep detection framework, which can boost target detection task with little annotation burden.}
\label{DetectVisualPOTD}
\end{figure*}

\begin{figure*}[htb!]
\centering
\subfigure[LSTD ]{\includegraphics[width=0.93\textwidth]{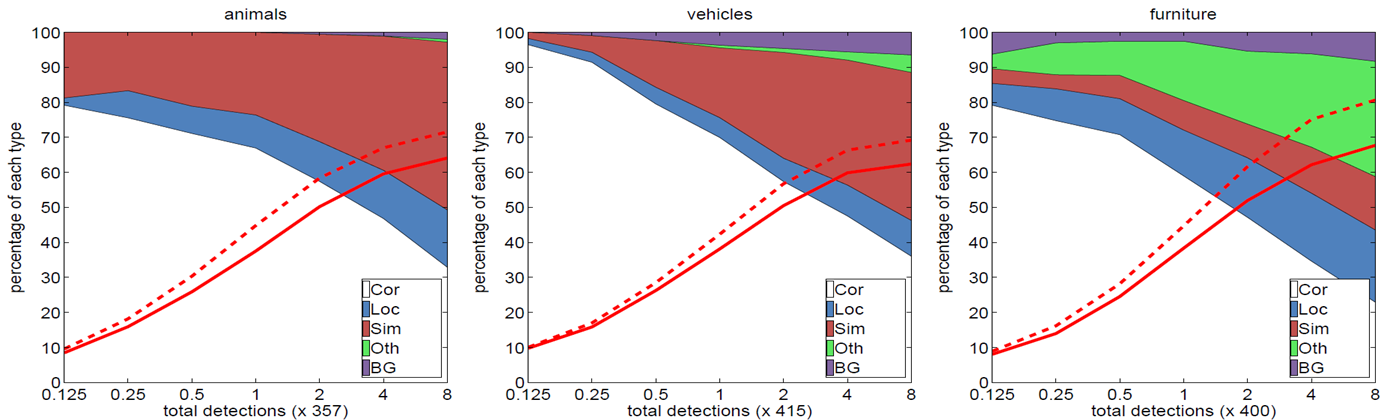}}
\subfigure[WSTD ]{\includegraphics[width=0.93\textwidth]{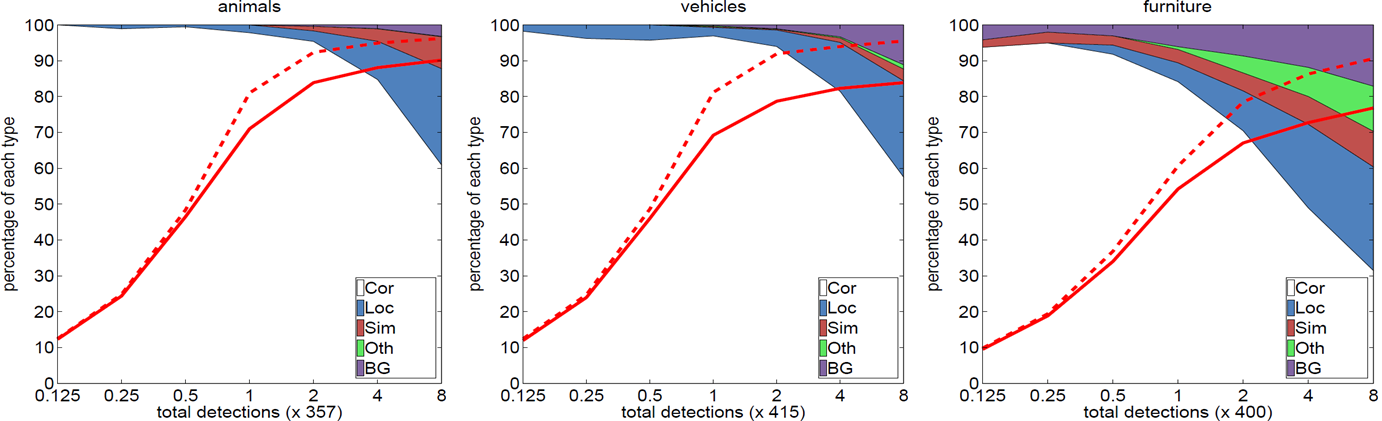}}
\subfigure[LSTD ]{\includegraphics[width=0.95\textwidth]{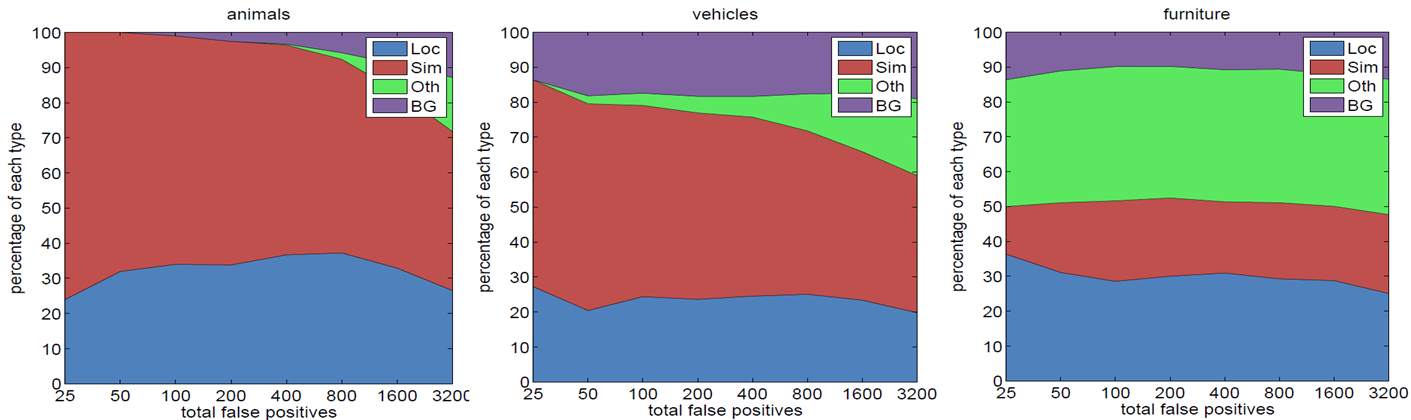}}
\subfigure[WSTD ]{\includegraphics[width=0.95\textwidth]{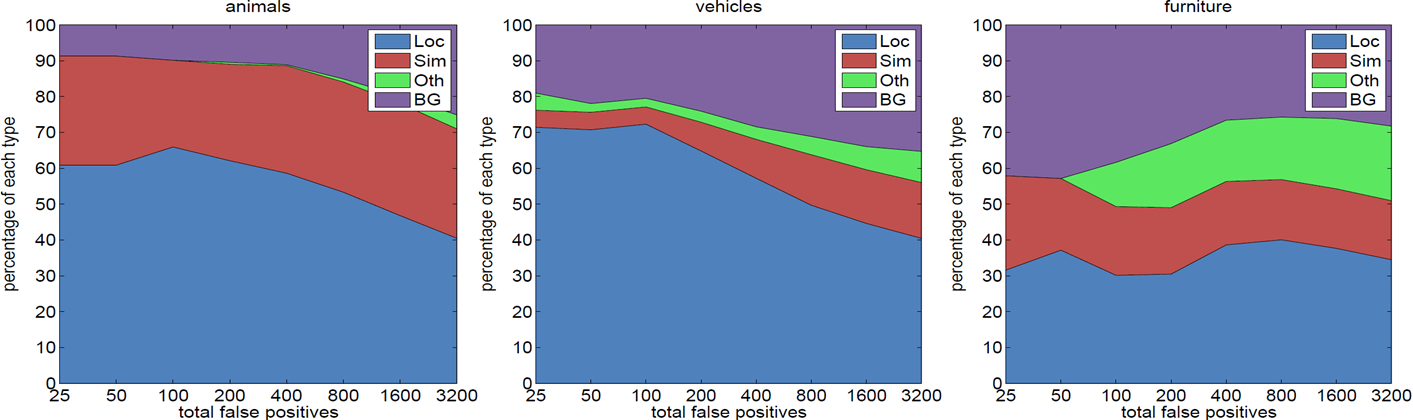}}
\caption{Error Mode Analysis.
The 1st-2nd row:
The cumulative fraction of detections which are correct (Cor) or false positive (i.e., Loc: poor localization, Sim/Oth/BG: confusion with similar categories/others/background).
The solid/dashed red line reflects the change of recall with 0.5/0.1 jaccard overlap as the number of detections increases.
The 3rd-4th row:
The distribution of top-ranked false positive types.}
\label{DetectError}
\end{figure*}

%


\textbf{Detection Visualization}.
We visualize two transfer detection stages of POTD in Fig. \ref{DetectVisualPOTD},
where one training image is fully annotated in the target domain (VOC2007).
First,
LSTD$_{(1shot)}$ can achieve a reasonably good performance,
via transferring object knowledge from source domain.
Second,
WSTD$_{(1shot)}$ can further generalize detection with weakly-annotated images.
Hence,
we can see that our approach can boost target-domain detection progressively and effectively but with little annotation burden.

\textbf{Error Mode Analysis}.
We compare LSTD$_{(1shot)}$ with WSTD$_{(1shot)}$,
according to error model analysis of VOC2007 in Fig. \ref{DetectError}.
First,
WSTD can largely reduce various error types of LSTD,
according to the 1st and 2nd row of Fig. \ref{DetectError}.
It shows that WSTD can further generalize target detector with weakly-annotated images.
Second,
one can see that in the 3rd and 4th rows of Fig. \ref{DetectError},
the distribution of error types for LSTD is different from the one for WSTD.
The main reason is that,
the LSTD stage is built on low-shot but fully-annotated images,
while
the WSTD stage is built on large-scale but weakly-annotated images.
In this case,
LSTD is largely confused by similar objects and/or others,
due to the lack of training images.
Alternatively,
WSTD is largely confused by poor localization and/or background,
due to the lack of object-level supervision.

\section{Conclusion}

In this paper,
we propose a novel progressive object transfer detection (POTD) framework.
First,
POTD effectively integrates various object supervision of different domains into a progressive detection procedure,
i.e.,
from source to target domains,
from large to few data,
from full to weak annotations.
Via this human-like learning,
POTD can boost a target detection task with little annotation burden.
Second,
each detection stage in POTD is efficiently designed with delicate transfer insights,
where LSTD is used as warm-up for WSTD generalization.
Finally,
we conduct extensive experiments to show that POTD outperforms other state-of-the-art methods.

\ifCLASSOPTIONcaptionsoff
  \newpage
\fi



\bibliographystyle{IEEEtran}
\bibliography{chen-egbib}

\begin{thebibliography}{10}
\providecommand{\url}[1]{#1}
\csname url@samestyle\endcsname
\providecommand{\newblock}{\relax}
\providecommand{\bibinfo}[2]{#2}
\providecommand{\BIBentrySTDinterwordspacing}{\spaceskip=0pt\relax}
\providecommand{\BIBentryALTinterwordstretchfactor}{4}
\providecommand{\BIBentryALTinterwordspacing}{\spaceskip=\fontdimen2\font plus
\BIBentryALTinterwordstretchfactor\fontdimen3\font minus
  \fontdimen4\font\relax}
\providecommand{\BIBforeignlanguage}[2]{{%
\expandafter\ifx\csname l@#1\endcsname\relax
\typeout{** WARNING: IEEEtran.bst: No hyphenation pattern has been}%
\typeout{** loaded for the language `#1'. Using the pattern for}%
\typeout{** the default language instead.}%
\else
\language=\csname l@#1\endcsname
\fi
#2}}
\providecommand{\BIBdecl}{\relax}
\BIBdecl

\bibitem{7410526}
R.~Girshick, ``{Fast R-CNN},'' in \emph{ICCV}, 2015.

\bibitem{ren2015faster}
S.~Ren, K.~He, R.~Girshick, and J.~Sun, ``{Faster R-CNN: Towards real-time
  object detection with region proposal networks},'' in \emph{NIPS}, 2015.

\bibitem{Liu2016SSDSS}
W.~Liu, D.~Anguelov, D.~Erhan, C.~Szegedy, S.~E. Reed, C.-Y. Fu, and A.~C.
  Berg, ``{SSD: Single Shot MultiBox Detector},'' in \emph{ECCV}, 2016.

\bibitem{he2017mask}
K.~He, G.~Gkioxari, P.~Doll{\'a}r, and R.~Girshick, ``{Mask R-CNN},'' in
  \emph{ICCV}, 2017.

\bibitem{8100028}
A.~Diba, V.~Sharma, A.~Pazandeh, H.~Pirsiavash, and L.~V. Gool, ``{Weakly
  Supervised Cascaded Convolutional Networks},'' in \emph{CVPR}, 2017, pp.
  5131--5139.

\bibitem{7780751}
D.~Li, J.~B. Huang, Y.~Li, S.~Wang, and M.~H. Yang, ``{Weakly Supervised Object
  Localization with Progressive Domain Adaptation},'' in \emph{CVPR}, 2016.

\bibitem{7780680}
H.~Bilen and A.~Vedaldi, ``{Weakly Supervised Deep Detection Networks},'' in
  \emph{CVPR}, 2016, pp. 2846--2854.

\bibitem{Lai2017SaliencyGE}
B.~Lai and X.~Gong, ``{Saliency Guided End-to-End Learning for Weakly
  Supervised Object Detection},'' in \emph{IJCAI}, 2017.

\bibitem{8099809}
P.~Tang, X.~Wang, X.~Bai, and W.~Liu, ``{Multiple Instance Detection Network
  with Online Instance Classifier Refinement},'' in \emph{CVPR}, 2017.

\bibitem{dong2017few}
X.~Dong, L.~Zheng, F.~Ma, Y.~Yang, and D.~Meng, ``{Few-shot Object
  Detection},'' \emph{arXiv:1706.08249}, 2017.

\bibitem{Hoffman2014LSDALS}
J.~Hoffman, S.~Guadarrama, E.~Tzeng, J.~Donahue, R.~B. Girshick, T.~Darrell,
  and K.~Saenko, ``{LSDA: Large Scale Detection Through Adaptation},'' in
  \emph{NIPS}, 2014.

\bibitem{7780602}
Y.~Tang, J.~Wang, B.~Gao, E.~Dellandréa, R.~Gaizauskas, and L.~Chen, ``{Large
  Scale Semi-Supervised Object Detection Using Visual and Semantic Knowledge
  Transfer},'' in \emph{CVPR}, 2016.

\bibitem{chen2018lstd}
H.~Chen, Y.~Wang, G.~Wang, and Y.~Qiao, ``{LSTD: A Low-Shot Transfer Detector
  for Object Detection},'' in \emph{AAAI}, 2018.

\bibitem{Liang2015TowardsCB}
X.~Liang, S.~Liu, Y.~Wei, L.~Liu, L.~Lin, and S.~Yan, ``{Towards Computational
  Baby Learning: A Weakly-Supervised Approach for Object Detection},''
  \emph{ICCV}, pp. 999--1007, 2015.

\bibitem{Dietterich1997SolvingTM}
T.~G. Dietterich, R.~H. Lathrop, and T.~Lozano-P{\'e}rez, ``{Solving the
  Multiple Instance Problem with Axis-Parallel Rectangles},'' \emph{Artif.
  Intell.}, vol.~89, pp. 31--71, 1997.

\bibitem{7420739}
R.~G. Cinbis, J.~Verbeek, and C.~Schmid, ``{Weakly Supervised Object
  Localization with Multi-Fold Multiple Instance Learning},'' \emph{PAMI},
  vol.~39, no.~1, pp. 189--203, Jan 2017.

\bibitem{7298711}
H.~Bilen, M.~Pedersoli, and T.~Tuytelaars, ``{Weakly supervised object
  detection with convex clustering},'' in \emph{CVPR}, 2015, pp. 1081--1089.

\bibitem{8099940}
Z.~Jie, Y.~Wei, X.~Jin, J.~Feng, and W.~Liu, ``{Deep Self-Taught Learning for
  Weakly Supervised Object Localization},'' in \emph{CVPR}, 2017, pp.
  4294--4302.

\bibitem{Kantorov2016ContextLocNetCD}
V.~Kantorov, M.~Oquab, M.~Cho, and I.~Laptev, ``{ContextLocNet: Context-Aware
  Deep Network Models for Weakly Supervised Localization},'' in \emph{ECCV},
  2016.

\bibitem{Uijlings2013SelectiveSF}
J.~R.~R. Uijlings, K.~E.~A. van~de Sande, T.~Gevers, and A.~W.~M. Smeulders,
  ``{Selective Search for Object Recognition},'' \emph{IJCV}, vol. 104, pp.
  154--171, 2013.

\bibitem{Zitnick2014EdgeBL}
C.~L. Zitnick and P.~Doll{\'a}r, ``{Edge Boxes: Locating Object Proposals from
  Edges},'' in \emph{ECCV}, 2014.

\bibitem{Renpami2016}
S.~Ren, K.~He, R.~Girshick, and J.~Sun, ``{Faster R-CNN: Towards Real-Time
  Object Detection with Region Proposal Networks},'' \emph{IEEE TPAMI}, 2016.

\bibitem{hinton2015distilling}
G.~Hinton, O.~Vinyals, and J.~Dean, ``{Distilling the knowledge in a neural
  network},'' \emph{arXiv preprint arXiv:1503.02531}, 2015.

\bibitem{Liueccv2016}
W.~Liu, D.~Anguelov, D.~Erhan, C.~Szegedy, S.~Reed, C.~Fu, and A.~C. Berg,
  ``{SSD: Single Shot MultiBox Detector},'' in \emph{ECCV}, 2016.

\bibitem{coco}
T.-Y. Lin, M.~Maire, S.~Belongie, J.~Hays, P.~Perona, D.~Ramanan,
  P.~Doll{\'{a}}r, and C.~L. Zitnick, ``{Microsoft {COCO}: Common objects in
  context},'' in \emph{ECCV}, 2014.

\bibitem{imagenet}
J.~Deng, W.~Dong, R.~Socher, L.-J. Li, K.~Li, and L.~Fei-Fei, ``{Imagenet: A
  large-scale hierarchical image database},'' in \emph{CVPR}, 2009.

\bibitem{voc}
M.~Everingham, L.~{Van Gool}, C.~K.~I. Williams, J.~Winn, and A.~Zisserman,
  ``{The Pascal Visual Object Classes {(VOC)} Challenge},'' \emph{IJCV}, 2010.

\bibitem{simonyan2014very}
K.~Simonyan and A.~Zisserman, ``{Very deep convolutional networks for
  large-scale image recognition},'' \emph{arXiv:1409.1556}, 2014.

\bibitem{Kingma2015}
D.~P. Kingma and J.~L. Ba, ``{Adam: a Method for Stochastic Optimization},'' in
  \emph{ICLR}, 2015.

\bibitem{jia2014caffe}
Y.~Jia, E.~Shelhamer, J.~Donahue, S.~Karayev, J.~Long, R.~Girshick,
  S.~Guadarrama, and T.~Darrell, ``{Caffe: Convolutional Architecture for Fast
  Feature Embedding},'' in \emph{arXiv preprint arXiv:1408.5093}, 2014.

\bibitem{bilen2016weaklyddn}
H.~Bilen and A.~Vedaldi, ``{Weakly supervised deep detection networks},'' in
  \emph{CVPR}, 2016.

\bibitem{diba2016weakly}
A.~Diba, V.~Sharma, A.~Pazandeh, H.~Pirsiavash, and L.~{Van Gool}, ``{Weakly
  Supervised Cascaded Convolutional Networks},'' \emph{arXiv:1611.08258}, 2016.

\bibitem{li2016weakly}
D.~Li, J.-B. Huang, Y.~Li, S.~Wang, and M.-H. Yang, ``{Weakly supervised object
  localization with progressive domain adaptation},'' in \emph{CVPR}, 2016.

\end{thebibliography}

\begin{IEEEbiography}[{\includegraphics[width=1in,height=1.25in,clip,keepaspectratio]{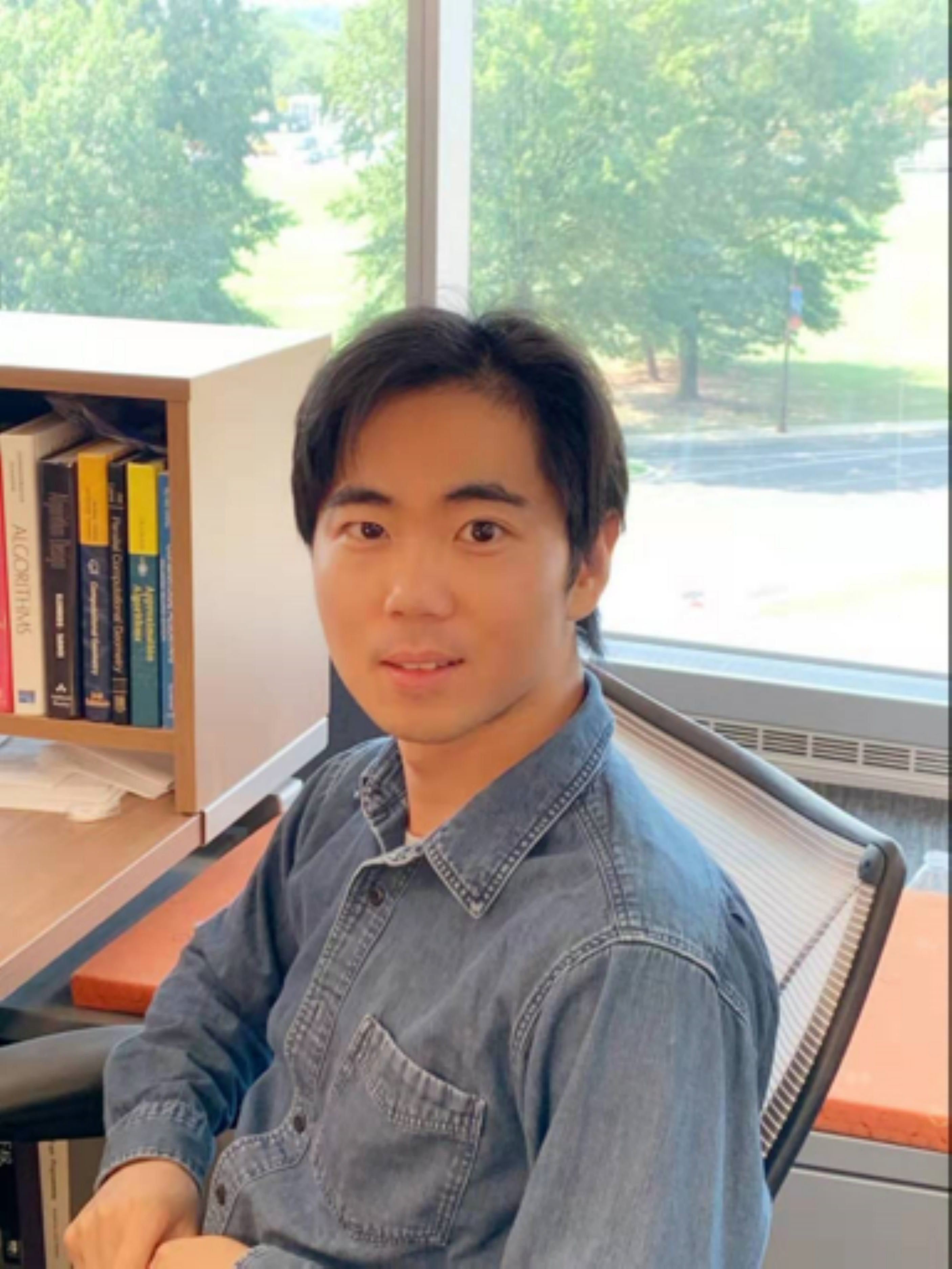}}]{Hao Chen}
received his B.S. degree in optical-electronic information engineering and M.S. degree in  pattern recognition and intelligent system from Huazhong University of Science and Technology (HUST), Wuhan, China, in 2015 and 2018, respectively.
He is currently a PhD student in computer science at University of Maryland, College Park.
His research interests include object detection, video understanding and deep learning.
\end{IEEEbiography}

\begin{IEEEbiography}[{\includegraphics[width=1in,height=1.25in,clip,keepaspectratio]{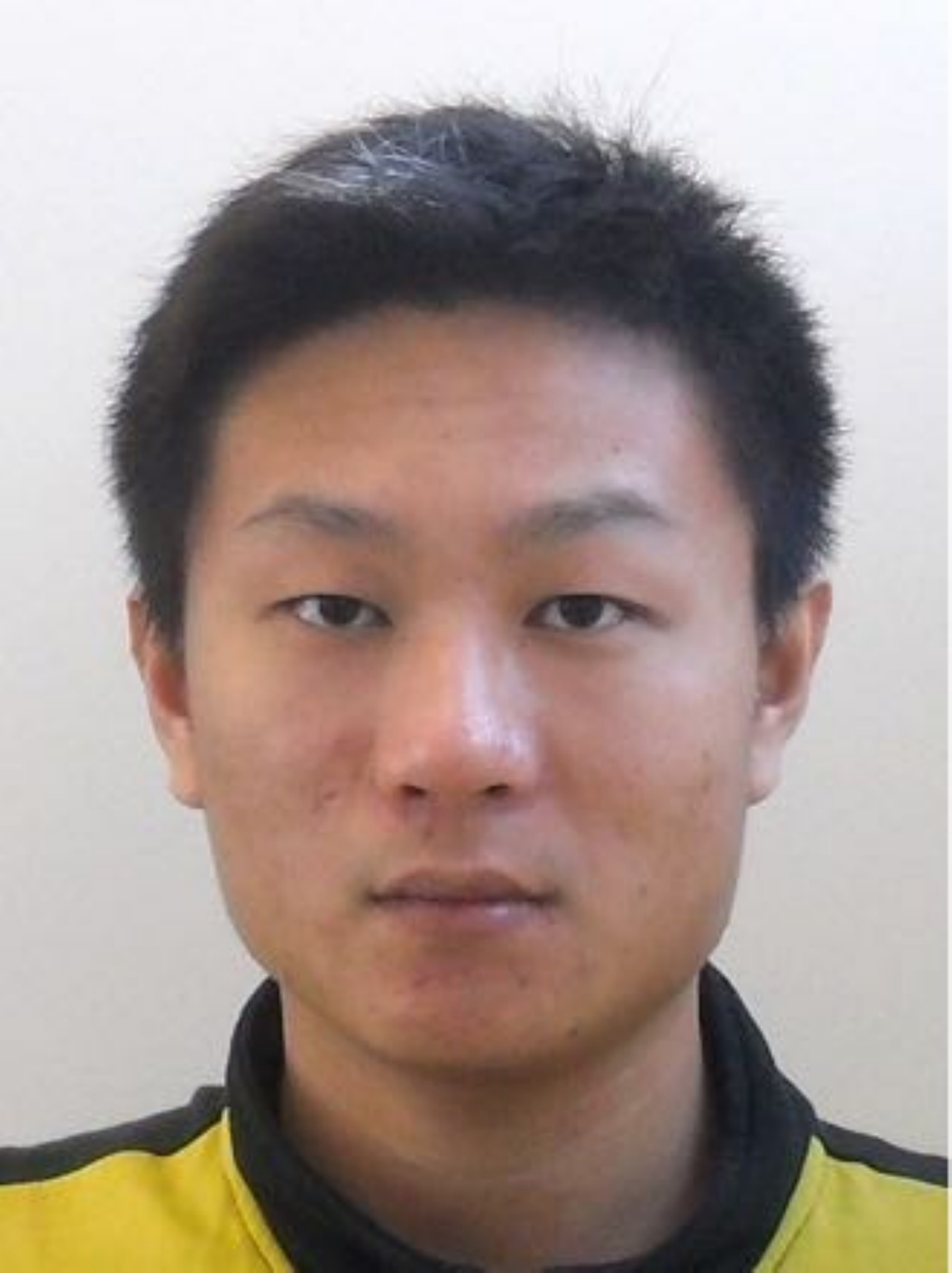}}]{Yali Wang}
received the Ph.D. degree in computer science from Laval University, Canada, in 2014.
He is currently an Associate Professor with Shenzhen Institutes of Advanced Technology, Chinese Academy of Sciences.
His research interests are deep learning and computer vision, machine learning and statistics.
\end{IEEEbiography}

\begin{IEEEbiography}[{\includegraphics[width=1in,height=1.25in,clip,keepaspectratio]{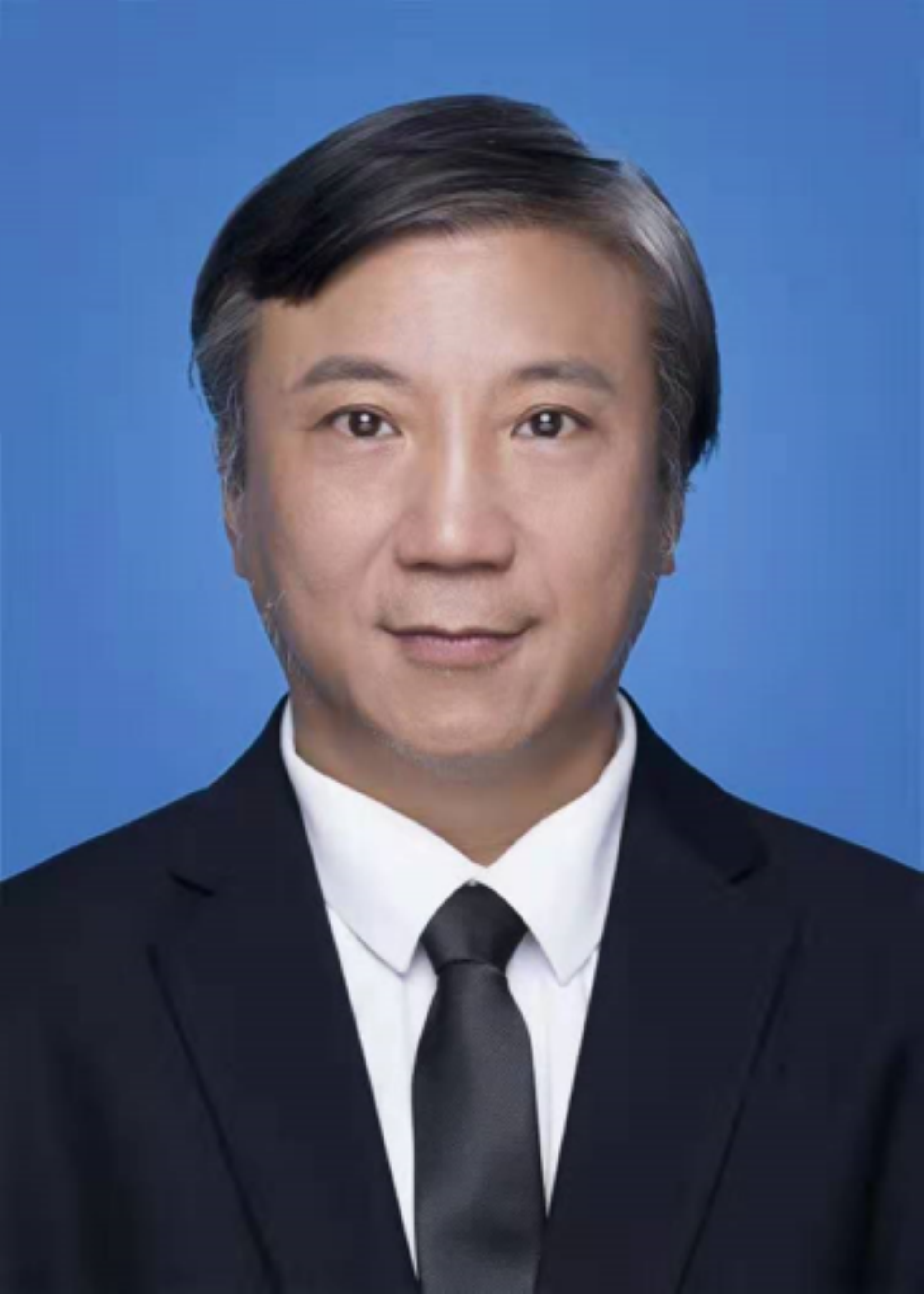}}]{Guoyou Wang}
received the B.S. degree in electronic engineering and the M.S. degree in pattern recognition and intelligent system from Huazhong University of Science and Technology, Wuhan, China, in 1988 and 1992, respectively.
He is currently a Professor with the Institute for Pattern Recognition and Artificial Intelligence, Huazhong University of Science and Technology.
His current research interests include image processing, image compression, pattern recognition, artificial intelligence, and machine learning.
\end{IEEEbiography}

\begin{IEEEbiography}[{\includegraphics[width=1in,height=1.25in,clip,keepaspectratio]{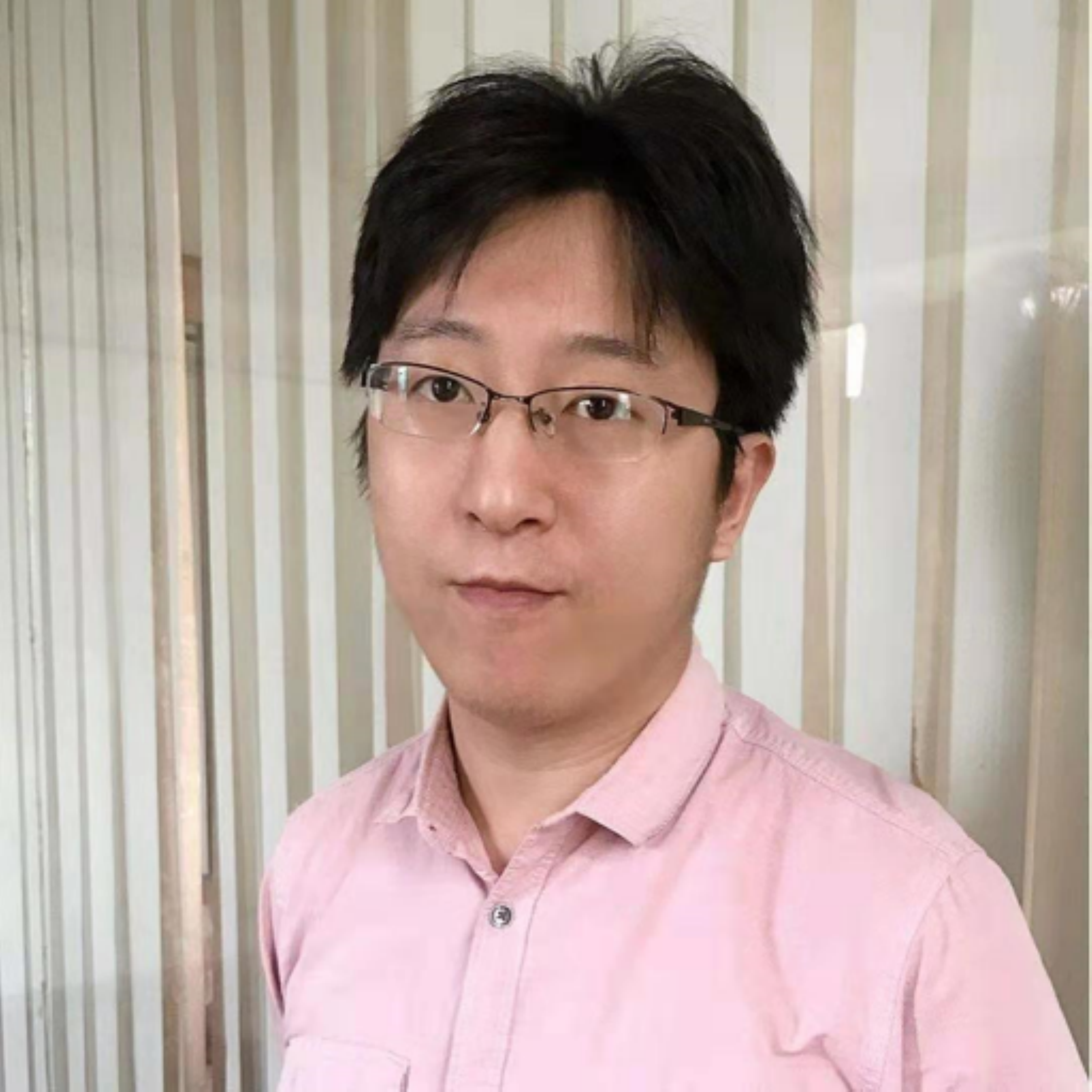}}]{Xiang Bai}
received his B.S., M.S., and Ph.D. degrees from Huazhong University of Science and Technology (HUST), Wuhan, China, in 2003, 2005, and 2009, respectively, all in
electronics and information engineering.
He is currently a Professor with the School of Electronic Information and Communications, HUST.
His research interests include object recognition, shape analysis, and OCR.
He received IAPR/ICDAR Young Investigator Award in 2019.
He is an associate editor for Pattern Recognition, Pattern Recognition Letters, and Frontiers of Computer Science.
\end{IEEEbiography}

\begin{IEEEbiography}[{\includegraphics[width=1in,height=1.25in,clip,keepaspectratio]{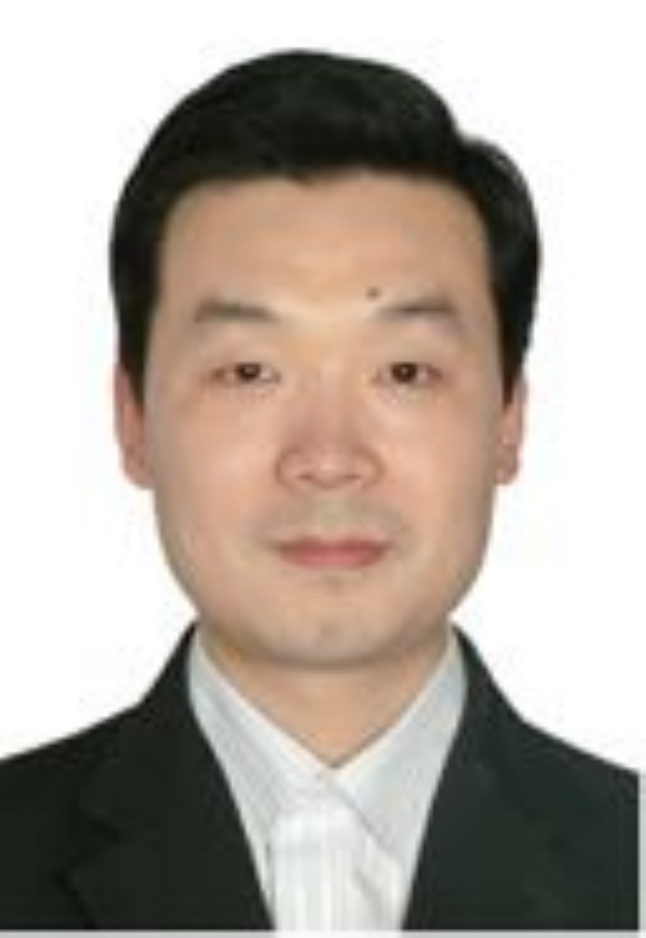}}]{Yu Qiao}
(SM'13) received the Ph.D. degree from The University of Electro-Communications, Japan, in 2006.
He was a JSPS Fellow and a Project Assistant Professor with The University of Tokyo from 2007 to 2010.
He is currently a Professor with Shenzhen Institutes of Advanced Technology, Chinese Academy of Sciences.
He has authored over 140 papers in journals and conference including, PAMI, IJCV, TIP, ICCV, CVPR, ECCV, AAAI.
His research interests include computer vision, deep learning, and intelligent robots.
He received the Lu Jiaxi Young Researcher Award from the Chinese Academy of Sciences in 2012.
He was the first runner-up at the ImageNet Large Scale Visual Recognition Challenge 2015 in scene recognition and the winner at the ActivityNet Large Scale Activity Recognition Challenge 2016 in video classification.
\end{IEEEbiography}

\end{document}